\documentclass[10pt,journal,letterpaper]{IEEEtran}
\usepackage[T1]{fontenc}
\usepackage{amsmath,amsfonts}
\usepackage{array}
\usepackage{booktabs}
\usepackage{pifont}
\usepackage{multirow}
\usepackage[table]{xcolor}
\usepackage{url}
\usepackage{graphicx}
\usepackage{stfloats}
\usepackage{cite}
\usepackage{microtype}
\usepackage{arydshln}

\definecolor{OursBlue}{RGB}{232,242,252}
\newcommand{\method}{DiFiT}
\definecolor{ConditionGreen}{RGB}{0,128,70}
\definecolor{ConditionRed}{RGB}{190,45,45}
\newcommand{\RequiredMark}{\textcolor{ConditionGreen}{\ding{51}}}
\newcommand{\NotRequiredMark}{\textcolor{ConditionRed}{\ding{55}}}

\usepackage[colorlinks=true,citecolor=blue,linkcolor=red,urlcolor=black]{hyperref}
\hypersetup{
  pdftitle={DiFiT: Distribution-Fidelity-guided Visual Token Pruning for Multimodal Large Language Models},
  pdfauthor={Jingyuan Zhang, Nihao Niu, Xinyang Lv, Beibei Han, Xiaoyun Wang, Yanning Hou, Ke Xu},
  pdfsubject={Visual token pruning},
  pdfkeywords={multimodal large language models, visual token pruning, efficient inference}
}

\usepackage{tikz}
\definecolor{lime}{HTML}{A6CE39}
\DeclareRobustCommand{\orcidicon}{%
    \begin{tikzpicture}
    \draw[lime, fill=lime] (0,0) 
    circle [radius=0.16] 
    node[white] {{\fontfamily{qag}\selectfont \tiny ID}};    \draw[white, fill=white] (-0.0625,0.095) 
    circle [radius=0.007];    \end{tikzpicture}
    \hspace{-2mm}}
\foreach \x in {A, ..., Z}{%
    \expandafter\xdef\csname orcid\x\endcsname{\noexpand\href{https://orcid.org/\csname orcidauthor\x\endcsname}{\noexpand\orcidicon}}
    }

\begin{document}
\bstctlcite{IEEEexample:BSTcontrol}

\title{DiFiT: Distribution-Fidelity-guided Visual Token Pruning for Multimodal Large Language Models}

\newcommand{\orcidauthorA}{0009-0005-4370-095X} 
\newcommand{\orcidauthorB}{0009-0001-2008-7583} 
\newcommand{\orcidauthorC}{0009-0006-4498-5173} 
\newcommand{\orcidauthorD}{0009-0009-6368-3121} 
\newcommand{\orcidauthorE}{0009-0005-4332-1677} 
\newcommand{\orcidauthorF}{0009-0002-9146-4494}  
\newcommand{\orcidauthorG}{0000-0001-6266-4257}  

\author{
\IEEEauthorblockN{
  Jingyuan~Zhang\orcidA{},
  Nihao~Niu\orcidB{}, 
  Xinyang~Lv\orcidC{},
  Beibei~Han\orcidD{},
  Xiaoyun~Wang\orcidE{},\\
  Yanning~Hou\orcidF{},
  \mbox{and Ke~Xu\orcidG{}, \IEEEmembership{Member,~IEEE}}
}
\thanks{Corresponding author: Xiaoyun Wang.}%
\thanks{Jingyuan Zhang, Nihao Niu, Xinyang Lv, Beibei Han, and Xiaoyun Wang
are with the School of Artificial Intelligence, Anhui University, Hefei 230601, China
(e-mail: \protect\nolinkurl{zhangjingyuan@stu.ahu.edu.cn}; \protect\nolinkurl{nhniu@stu.ahu.edu.cn};
\protect\nolinkurl{e24301262@stu.ahu.edu.cn}; \protect\nolinkurl{isxiaohanw@gmail.com}; \protect\nolinkurl{wangxiaoyun@stu.ahu.edu.cn}).}%
\thanks{Yanning Hou is with the College of Intelligence Science and Technology,
National University of Defense Technology, Changsha 410073, China
(e-mail: \protect\nolinkurl{yanning_hou@nudt.edu.cn}).}%
\thanks{Ke Xu is with the State Key Laboratory of Opto-Electronic Information
Acquisition and Protection Technology, Anhui University, Hefei 230601, China
(e-mail: \protect\nolinkurl{xuke@ahu.edu.cn}).}
\thanks{This work was supported in part by the National Natural Science Foundation of China (No. 62576001, No. 62206003).}
}

\maketitle

\setlength{\emergencystretch}{1.2em}

\begin{abstract}
Visual token pruning reduces the inference cost of multimodal large language models, but answer accuracy only partly reflects its effect. Pruning can change the probabilities of alternative response tokens even when greedy answers remain correct and unchanged. We compare full and pruned response distributions under shared reference prefixes. Average predictive drift grows with compression on fixed correctly answered samples. Separate analyses associate larger drift with more confidence-decision disagreements and sampling failures. We propose \method{} (Distribution-Fidelity-guided Token pruning) to learn discrete selection from reference answers and full-token response distributions. Dual-View Distillation matches probabilities across the vocabulary, while answer supervision supports task quality. A learned scorer and differentiable Top-$K$ learn which visual evidence to retain. A multi-budget mask objective supports different retention ratios through one shared ranking over a frozen model. On Qwen2.5-VL-7B, \method{} reduces sampling-induced negative flips by 30.3\% relative to FastV at 10\% retention on four-task greedy-correct controls. It preserves 98.74\% of baseline performance on MME at 30\% retention.
\end{abstract}

\begin{IEEEkeywords}
Multimodal large language models, visual token pruning, efficient inference.
\end{IEEEkeywords}

\section{Introduction}\label{sec:intro}

\IEEEPARstart{M}{ultimodal} large language models (MLLMs) encode images and videos as visual tokens for language generation~\cite{liu2023llava, wang2025internvl3_5, bai2025qwen25vl, li2024llavaonevision}. Detailed inputs create longer sequences and increase memory use and prefill latency~\cite{li2024llavanext-ablations}. Visual token pruning and compression lower these costs by retaining or combining fewer tokens~\cite{chen2024fastv, yang2025visionzip, shang2025prumerge, Ye_2025_CVPR}. Methods are usually compared by the task performance retained at a given budget. These scores show whether answers stay correct but reveal little about changes in the probabilities assigned to possible responses.

Existing methods select tokens through importance scoring, coverage, and learning. Attention estimates which tokens matter to the model~\cite{chen2024fastv, zhang2025sparsevlm, arif2025hired}. Redundancy reduction and diversity-aware selection preserve complementary information~\cite{yang2025visionzip, alvar2025divprune, bolya2023tome, zhang2025cdpruner}. Learned selectors adapt to the input and task~\cite{huang2025dynamicllava, liu2024mustdrop, ye2025atpllava, zhu2026visionselector}. Attention sinks, position effects, and incomplete coverage can still affect selection~\cite{xiao2024streamingllm, gu2025attentionsink, kang2025visualattnsink, endo2025feather, holov2025, zhang2025vispruner}. These methods address how to select tokens. We ask what the retained subset should preserve about the full model's predictions beyond the final answer.

Consider full-token and pruned models that produce the same greedy answer. At each response position, both give the same token the highest probability. One distribution may concentrate on that token, while the other assigns more weight to alternatives. A confidence policy can then accept different responses~\cite{geifman2019selectivenet, elyaniv2010selective}, and stochastic decoding may sample different continuations~\cite{wang2023selfconsistency, holtzman2020degeneration}. We call the vocabulary probabilities at each response position the \emph{predictive distribution}, and their change after pruning \emph{predictive drift}. Greedy answer agreement alone cannot reveal these changes in response probabilities.

\begin{figure*}[t]
  \centering
  \includegraphics[width=0.318\textwidth]{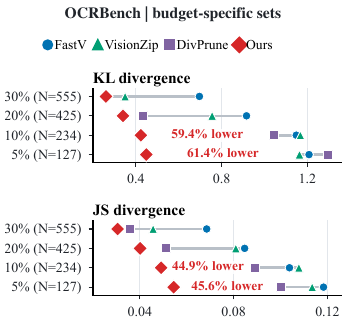}\hfill
  \includegraphics[width=0.318\textwidth]{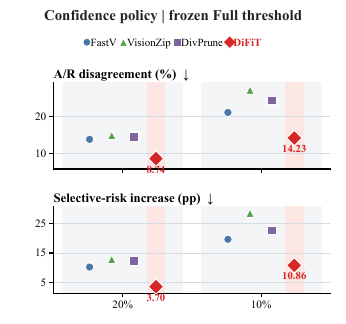}\hfill
  \includegraphics[width=0.318\textwidth]{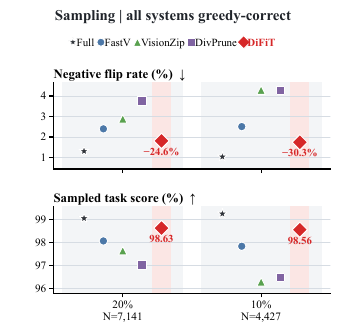}
  \caption{Probability drift and behavior changes beyond greedy answers. Left: KL/JS drift on OCRBench common-correct sets at each budget, a broad control. Middle: four-task accept/reject disagreement and selective-risk increase when reusing a full-model threshold set for 90\% coverage. Right: negative flips and sampled score on sets all systems answer correctly under greedy decoding ($N=7{,}141$ at 20\%; $N=4{,}427$ at 10\%). Pale-red bands identify \method{} and are not confidence intervals. Figure~\ref{fig:same_output_drift} and Table~\ref{tab:drift_failure_regression} provide stricter output controls and adjusted drift--behavior associations.}
  \label{fig:motivation}
\end{figure*}

We test this possibility under shared reference prefixes, which keep text history fixed. On fixed samples with identical correct answers, average drift grows with compression for every evaluated pruning method (Table~\ref{tab:strict_fidelity}(B)). Figure~\ref{fig:motivation} summarizes drift and behavior under broader correctness controls. Separate analyses link larger drift to more confidence-decision disagreements and sampling failures (Fig.~\ref{fig:drift_failure} and Table~\ref{tab:drift_failure_regression}). These associations persist after adjustment for full-model confidence, first-token prediction margin, task, and token budget. These results link response probabilities to changes in confidence decisions and sampling beyond greedy correctness.

Reference answers and full-token probabilities provide complementary supervision. Reference answers identify which response tokens to reward. The full-token distribution describes how the model divides probability among competing outputs~\cite{hinton2015distilling, li2025fcotvl}. Cross-entropy changes the entire distribution but does not directly match alternative-token probabilities to the full model. Different token subsets can therefore receive similar answer supervision while preserving different distributions. Teacher probabilities provide this additional selection target. The full model may still be wrong or poorly calibrated~\cite{guo2017calibration, tan2026calibration}. Combining answer supervision with full-token distribution matching supports both task quality and predictive fidelity.

We propose \method{} (Distribution-Fidelity-guided Token pruning) to learn visual-token selection from this combined objective. Dual-View Distillation (DVD) matches response distributions under shared reference prefixes. The Dual-Perception Scorer (DPS) and differentiable Top-$K$ make discrete selection learnable~\cite{ahle2023difftopk}. DPS scores visual importance and can incorporate text relevance, and prediction-loss gradients train these scores. Matryoshka Budget Loss (MBL) supervises soft masks across budgets using hard selection and larger-budget guidance. This shared ranking supports different token retention ratios with one trained scorer. The vision encoder, projector, and language model stay frozen. Inference keeps the highest-scoring tokens for one pruned forward pass compatible with efficient attention kernels~\cite{dao2022flashattention}.

We evaluate task performance across three backbones, eight image-language benchmarks, and MVBench, training a separate scorer for each backbone. Drift and behavior analyses use Qwen2.5-VL-7B, with behavior tests covering four tasks and two budgets. The main contributions of this work are as follows.
\begin{itemize}
  \item We quantify predictive drift hidden by unchanged greedy answers. Under shared reference prefixes and fixed correct output-matched samples, average drift grows with stronger compression for every evaluated pruning method.
  \item We identify an association between predictive drift and behavior. Larger drift accompanies more confidence-decision disagreements and sampling failures. This association persists after adjustment for full-model confidence, prediction margin, task, and budget.
  \item We introduce \method{} to learn discrete visual-token selection from reference answers and full-token response distributions. Learned scoring and multi-budget mask supervision support different retention ratios over a frozen MLLM. Task and behavior evaluations show improved performance retention and consistency after pruning.
\end{itemize}

Section~\ref{sec:related} reviews visual token compression and knowledge distillation. Section~\ref{sec:method} presents \method{} and its training objectives. Section~\ref{sec:experiments} describes the setup, task results, behavior analyses, ablations, and limitations. Section~\ref{sec:conclusion} presents the conclusion.

\section{Related Work}\label{sec:related}

\subsection{Visual Token Compression}

Visual token compression seeks a smaller input that retains the evidence needed for a response. Attention-based methods rank individual tokens by importance~\cite{chen2024fastv, zhang2025sparsevlm, arif2025hired}. Query guidance and explainability refine relevance estimates~\cite{gao2026quietprune, lei2026taskrelated}. Such scores can be affected by attention sinks and positional effects~\cite{xiao2024streamingllm, gu2025attentionsink, kang2025visualattnsink, endo2025feather, holov2025, zhang2025vispruner}. Several highly ranked tokens may describe the same region. Other useful evidence can then remain uncovered. Redundancy reduction and merging address repeated information~\cite{yang2025visionzip, shang2025prumerge, bolya2023tome, wen2025dart, yu2026visiontrim}. Diversity- and region-aware methods focus on complementary information and spatial coverage~\cite{alvar2025divprune, zhang2025cdpruner, fang2026prunesid, deng2025scope, liang2026pyramid}. Video compression extends pruning and merging across frames~\cite{su2025stpm}. These criteria assess the retained set through visual redundancy, diversity, or coverage. They do not directly specify the response probabilities that it should preserve. Learned compression and input-dependent sparsification make selection adaptable to the input and task~\cite{Ye_2025_CVPR, huang2025dynamicllava, cao2024madtp, liu2024mustdrop, ye2025atpllava}. VisionSelector uses a lightweight scorer and differentiable Top-$K$ for end-to-end token selection~\cite{zhu2026visionselector, ahle2023difftopk}. Its objective combines task cross-entropy with consistency between soft and hard selection masks. It does not explicitly match the pruned response distribution to the full-token model. Preservation objectives also differ. ApET uses reconstruction error~\cite{ma2026apet}, DiViCo estimates information loss~\cite{wang2026divico}, and IF-Prune uses a small VLM's variational information bottleneck~\cite{sun2026ifprune}. OTPrune aligns full and pruned visual-token distributions through optimal transport, targeting diversity and representativeness~\cite{chen2026otprune}. These approaches assess retained information through visual representations or information proxies. \method{} places its preservation target at the model output: the vocabulary probabilities at response positions. It uses these response probabilities and reference answers to supervise which existing tokens should remain in the final pruned visual input.

\subsection{Knowledge Distillation and Predictive Fidelity}

Knowledge distillation transfers information from teacher predictions beyond hard labels~\cite{hinton2015distilling}. Soft probabilities describe the relative support for competing outputs. Their usefulness depends on the teacher and learning setting~\cite{muller2019when, furlanello2018born, zhou2021rethinking, menon2021statistical, cho2019efficacy}. Distillation has been applied to multimodal model compression~\cite{cai2025llavakd, wang2023efficientvlm} and token reduction~\cite{jie2024tocom}. These methods differ in what is compressed, which predictions are matched, and which parameters receive updates. Output-probability changes have also been used to characterize token information across model layers~\cite{wang2026informationhorizon}. FCoT-VL is closely related. It combines output-distribution KL with answer supervision for visual compression~\cite{li2025fcotvl}. Its self-distillation stage learns a compression module and projector to construct a compact visual representation. \method{} instead applies predictive supervision to discrete token retention. It updates only the selector while keeping the vision encoder, projector, and language model frozen. The distinction lies in how predictive supervision is used: learning compressed representations or learning which existing tokens to retain. We further examine predictive drift that remains hidden when full-token and pruned models produce the same greedy answers.

\section{Method}\label{sec:method}

Given a query and visual tokens $V$, our goal is to retain a subset $V_b$ at ratio $b$ that preserves answer quality and response probabilities. We learn the ranking through a softly weighted training view, supervised by DVD and the reference answer, then use hard token selection at inference.

\begin{figure*}[t]
    \centering
    \includegraphics[width=\textwidth]{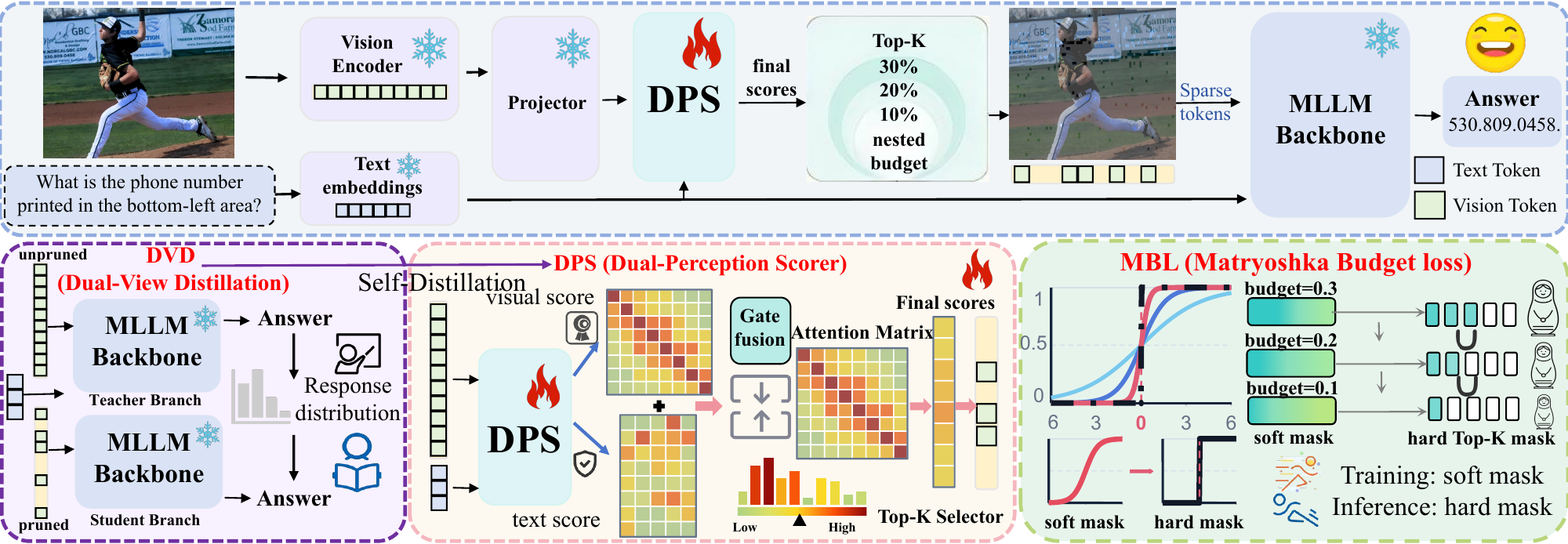}
    \caption{Overview of \method{}. DVD matches the softly weighted student view to full-token response distributions. DPS scores visual importance and text relevance. MBL aligns soft masks with hard selection across budgets. CE and DVD supervise the 30\% student path; MBL supervises masks at 30\%, 20\%, and 10\%. Inference gathers hard Top-$K$ tokens into a shortened sequence for generation.}
    \label{fig:framework}
\end{figure*}

Figure~\ref{fig:framework} connects prediction supervision to token selection through DPS, differentiable Top-$K$, and MBL. The formulation below describes our main Qwen2.5-VL-7B setting. Training updates the scorer while keeping the vision encoder, projector, and language model frozen.

\subsection{Dual-View Distillation for Predictive Fidelity}\label{sec:dvd}

DVD trains the selector to preserve full-token response probabilities~\cite{hinton2015distilling, muller2019when, zhou2021rethinking}. The teacher distribution indicates which alternative tokens the model considers plausible, providing supervision beyond the reference answer.

During training, the frozen backbone $f_\theta$ evaluates the full visual sequence $V=\{v_1,\dots,v_N\}$ and a softly weighted view:
\begin{equation}
z^{\mathrm{full}} = f_\theta(V,t), \qquad
z^{\mathrm{soft}} = f_\theta(\widetilde V_b,t),
\end{equation}
where $\widetilde V_b=\{M_{b,i}v_i\}_{i=1}^{N}$ uses the budget-constrained mask from Section~\ref{sec:dps}. Both views contain all $N$ visual positions and the same teacher-forced text $t$; only the visual features are weighted differently.

Let $\tilde z^{\mathrm{full}}$ and $\tilde z^{\mathrm{soft}}$ denote shifted logits, with $m_i\in\{0,1\}$ selecting supervised response positions. The teacher branch runs without gradients. In the student branch, gradients pass through the frozen language model and soft mask to the scorer.

Temperature scaling exposes this information by softening the output distributions. At each supervised response position, we divide both teacher and student logits by $\tau>1$ before applying softmax:
\begin{equation}
\begin{aligned}
P_i^{\mathrm{full}}&=\mathrm{softmax}\!\left(\tilde{z}_i^{\mathrm{full}}/\tau\right),\\
P_i^{\mathrm{soft}}&=\mathrm{softmax}\!\left(\tilde{z}_i^{\mathrm{soft}}/\tau\right).
\end{aligned}
\end{equation}
We then average the teacher-to-student KL divergence over the supervised response positions:
\begin{equation}
\mathcal{L}_{\mathrm{KL}}=\frac{\tau^2}{\sum_i m_i}\sum_i m_i \, D_{\mathrm{KL}}\!\left(P_i^{\mathrm{full}} \,\|\, P_i^{\mathrm{soft}}\right),
\end{equation}
The $\tau^2$ factor compensates for the gradient scaling introduced by the temperature~\cite{hinton2015distilling}. We set $\tau=2$; the most effective temperature can vary across tasks~\cite{cho2019efficacy, li2023ctkd}.

\subsection{Dual-Perception Scoring and Differentiable Selection}\label{sec:dps}

\label{sec:difftopk}

DPS learns token rankings from prediction supervision. In text-aware scoring, it combines visual importance with relevance to the available input tokens. Let $\mathbf{X}\in\mathbb{R}^{B\times N\times D}$ denote projected visual features, with batch size $B$, token count $N$, and embedding dimension $D$. In compact notation, the learnable query and key projections are
\begin{equation}
\mathbf{Q}=\mathbf{X}\mathbf{W}_q, \qquad
\mathbf{K}=\mathbf{X}\mathbf{W}_k,
\end{equation}
where $\mathbf W_q,\mathbf W_k\in\mathbb R^{D\times d}$.

The visual branch follows the row-averaged query--key scoring used by VisionSelector~\cite{zhu2026visionselector}. We express this score equivalently using the mean key $\bar{\mathbf{k}}=N^{-1}\sum_j\mathbf{k}_j$:
\begin{equation}
s_i^{\mathrm{vis}}=\frac{\mathbf{q}_i^{\top}\bar{\mathbf{k}}}{\sqrt{d}}.
\end{equation}
The implementation forms $\mathbf Q\mathbf K^\top/\sqrt d$ and averages its rows; the mean-key expression above is algebraically equivalent.

The text-aware branch compares $\mathbf X$ with the available input-token embeddings $\mathbf E$ in the shared LLM embedding space:
\begin{equation}
\mathbf{A}^{\mathrm{txt}}=\bar{\mathbf{X}}\,\bar{\mathbf{E}}^{\top}, \qquad
s_i^{\mathrm{txt}}=\max_j\,(\mathbf{A}^{\mathrm{txt}})_{ij},
\end{equation}
where bars denote $\ell_2$ normalization. Max pooling assigns each visual token its highest similarity to an input-token embedding.

A learnable scalar gate $\alpha$ combines the visual and text scores. We initialize it to zero:
\begin{equation}\label{eq:gated_fusion}
  s_i = s^{\mathrm{vis}}_i + \tanh\!\left( \alpha \right) \cdot s^{\mathrm{txt}}_i, \quad \alpha \in \mathbb{R}.
\end{equation}
The zero gate initially uses visual scores alone, while $\tanh(\alpha)\in[-1,1]$ bounds the text contribution. We initialize the entries of $\mathbf W_q$ and $\mathbf W_k$ from a Gaussian distribution with mean zero and standard deviation $10^{-4}$. For $L$ input-token embeddings, the implemented scorer costs $\mathcal O(NDd+N^2d+NLD)$, including pairwise visual scores.

Hard Top-$K$ selects discrete token indices, so it does not directly provide gradients for the scores. Continuous approximations include optimal-transport-based selection~\cite{xie2020difftopk}. We use a budget-constrained sigmoid mask~\cite{ahle2023difftopk}: for scores $\mathbf{s}$ and selection count $k$, the weights $\mathbf{M}\in[0,1]^{B\times N}$ satisfy
\begin{equation}
M_i=\sigma(s_i+\lambda),
\qquad
\sum_{i=1}^{N}M_i=k,
\label{eq:difftopk_forward}
\end{equation}
We find the offset $\lambda$ by bisection, preserving the score order while enforcing the budget. Because the weights must sum to $k$, increasing one token's weight reduces the weight available to the others. The scorer therefore learns to allocate a fixed budget across tokens.

The budget constraint also couples the gradients. Writing $v_i=M_i(1-M_i)$ and $g_i=\partial\mathcal{L}/\partial M_i$, implicit differentiation gives
\begin{equation}
\frac{\partial\mathcal{L}}{\partial s_i}
=
v_i\!\left(
 g_i-
 \frac{\sum_{j=1}^{N}g_jv_j}
      {\sum_{j=1}^{N}v_j}
\right),
\label{eq:difftopk_backward}
\end{equation}
The shared correction accounts for the fixed total mask weight when updating the scores.

During training, the student uses $M_{b,i}v_i$ at every visual position. This continuous weighting carries prediction-loss gradients to the scores without shortening the sequence. Hard Top-$K$ supplies auxiliary targets for MBL and selects the tokens physically retained at inference.

\subsection{Matryoshka Budget Loss}\label{sec:mbl}

Matryoshka Representation Learning trains nested embedding dimensions~\cite{kusupati2022matryoshka}. We use the same principle to train a shared token ranking at several budgets. Its hard Top-$K$ sets are already nested: tokens selected at 10\% also appear at 20\% and 30\%. MBL supervises the corresponding soft masks using both hard selection and guidance from the next larger budget.

For example, the 20\% mask learns from its own hard target of $\lfloor0.2N\rfloor$ tokens and from the 30\% soft mask; the 10\% mask similarly receives guidance from 20\%. The hard target encourages a clear selection boundary, while the larger-budget mask provides graded preferences among tokens. Combining the two trains the soft masks to allocate progressively smaller budgets using one ranking.

For training budget $b$, let $k_b=\lfloor bN\rfloor$. The differentiable rule in Sec.~\ref{sec:difftopk} maps the scores $s\in\mathbb R^N$ to a soft mask:
\begin{equation}
M_b=\mathcal{R}(s,k_b),
\qquad
\sum_{i=1}^{N}M_{b,i}=k_b,
\end{equation}
Here $M_b\in[0,1]^N$ is the soft mask. We also form a binary target from hard Top-$K$ selection:
\begin{equation}
M_b^{\mathrm{hard}}=\mathrm{TopK}(s,k_b),
\end{equation}
where $\mathrm{TopK}(\cdot)$ assigns one to the selected entries and zero to the others.

At the largest budget $b_0$, we train the soft mask against this hard target:
\begin{equation}
\mathcal{L}_0
=\mathrm{BCE}(M_{b_0},M_{b_0}^{\mathrm{hard}}).
\end{equation}

At each smaller budget $b_k$, we obtain a second target by scaling the soft mask from the next larger budget $b_{k-1}$:
\begin{equation}
p_{b_k}^{\mathrm{target}}
=\operatorname{sg}(M_{b_{k-1}})\cdot\frac{b_k}{b_{k-1}}.
\end{equation}
The stop-gradient operator $\operatorname{sg}(\cdot)$ keeps the larger-budget mask fixed when it serves as a target. Scaling by $b_k/b_{k-1}$ preserves its relative token preferences, while the student mask remains constrained to sum to $k_{b_k}$. We combine this target with the smaller budget's hard selection target:
\begin{equation}
\mathcal{L}_k
=\gamma\,\mathrm{BCE}(M_{b_k},M_{b_k}^{\mathrm{hard}})
+(1-\gamma)\,\mathrm{BCE}(M_{b_k},p_{b_k}^{\mathrm{target}}),
\end{equation}
The coefficient $\gamma$ controls the balance between hard selection and larger-budget guidance at the smaller budgets.

We average BCE over token entries and then over the $K$ budget levels:
\begin{equation}
  \mathcal{L}_{\mathrm{MBL}} = \frac{1}{K}\sum_{k=0}^{K-1} \mathcal{L}_k.
\end{equation}
We use $K=3$ budgets, $\{0.3,0.2,0.1\}$, and set $\gamma=0.5$. All masks are computed from the same scores, so one scorer supports the three retention ratios.

\begin{table*}[t]
\centering
\footnotesize
\caption{Image-language results on Qwen2.5-VL-7B. Avg. is the mean of eight task scores divided by their full-token scores (\%). Bold and underlined values mark the best and second-best pruning results; $\dagger$ marks training-based methods.}
\label{tab:comparison_7b}
\setlength{\tabcolsep}{4pt}
\renewcommand{\arraystretch}{1.00}
\resizebox{\textwidth}{!}{%
\begin{tabular}{lccccccccc}
\toprule
\multirow{2}{*}{\textbf{Method}}
& \textbf{DocVQA}
& \textbf{TextVQA}
& \textbf{OCRBench}
& \textbf{ScienceQA}
& \textbf{AI2D}
& \textbf{MMMU}
& \textbf{MME}
& \textbf{POPE}
& \multirow{2}{*}{\textbf{Avg.}} \\
& ANLS & Acc. & Acc. & EM & Acc. & Acc. & Score & F1 & \\
\midrule

Qwen2.5-VL-7B
& \textit{94.78}
& \textit{83.67}
& \textit{83.40}
& \textit{87.26}
& \textit{93.65}
& \textit{50.78}
& \textit{2342.15}
& \textit{86.19}
& \textit{100.00\%} \\

\midrule
\multicolumn{10}{c}{%
\cellcolor{gray!8}\textit{Retain 30\% Tokens (70\% Compression Ratio)}} \\
\midrule

FastV (ECCV'2024)~\cite{chen2024fastv}
& 84.01
& \textbf{80.22}
& 68.70
& 83.06
& 86.92
& 49.44
& 2263.58
& 80.47
& 92.78\% \\

LLaVA-PruMerge (ICCV'2025)~\cite{shang2025prumerge}
& 73.95
& 73.71
& 64.80
& 85.22
& 82.77
& 47.67
& 2239.64
& 83.69
& 89.56\% \\

VisionZip (CVPR'2025)~\cite{yang2025visionzip}
& 86.11
& 77.30
& 71.10
& \textbf{86.61}
& 87.86
& 49.44
& 2276.04
& 84.73
& 94.30\% \\

DivPrune (CVPR'2025)~\cite{alvar2025divprune}
& 82.51
& 78.52
& 72.00
& 86.01
& 88.28
& 48.33
& 2224.06
& 84.68
& 93.56\% \\

VisionTrim (ICLR'2026)~\cite{yu2026visiontrim}
& 69.74
& 61.28
& 69.60
& 84.23
& 82.42
& \textbf{50.78}
& 2296.97
& 82.16
& 88.53\% \\

PruneSiD (ICLR'2026)~\cite{fang2026prunesid}
& 77.04
& 71.21
& 62.20
& 83.74
& 80.12
& 48.22
& 2186.94
& \textbf{86.12}
& 88.84\% \\

\hdashline

Dynamic-LLaVA$^\dagger$ (ICLR'2025)~\cite{huang2025dynamicllava}
& 86.32
& 73.56
& 75.00
& 78.78
& 83.29
& 41.78
& 2180.42
& 80.87
& 89.67\% \\

VisionSelector$^\dagger$ (ACM MM'2026)~\cite{zhu2026visionselector}
& \underline{91.20}
& 79.37
& \underline{78.90}
& 86.37
& \textbf{92.10}
& \underline{49.56}
& \textbf{2341.31}
& \underline{85.99}
& \underline{97.54\%} \\

\rowcolor{OursBlue}
\textbf{\method{}}$^\dagger$
& \textbf{93.03}
& \underline{80.07}
& \textbf{79.50}
& \underline{86.42}
& \underline{91.97}
& \underline{49.56}
& \underline{2312.55}
& 85.59
& \textbf{97.76\%} \\

\midrule
\multicolumn{10}{c}{%
\cellcolor{gray!8}\textit{Retain 20\% Tokens (80\% Compression Ratio)}} \\
\midrule

FastV (ECCV'2024)~\cite{chen2024fastv}
& 75.99
& \underline{78.01}
& 59.70
& 82.75
& 82.35
& \underline{49.00}
& 2152.74
& 76.12
& 88.06\% \\

LLaVA-PruMerge (ICCV'2025)~\cite{shang2025prumerge}
& 61.09
& 68.72
& 56.20
& 83.79
& 77.62
& 46.11
& 2219.30
& 81.74
& 84.16\% \\

VisionZip (CVPR'2025)~\cite{yang2025visionzip}
& 74.75
& 72.03
& 59.10
& 84.68
& 82.32
& 47.00
& 2168.86
& 83.23
& 87.81\% \\

DivPrune (CVPR'2025)~\cite{alvar2025divprune}
& 73.81
& 73.86
& 64.80
& 84.33
& 82.29
& 46.33
& 2198.82
& 83.55
& 88.80\% \\

VisionTrim (ICLR'2026)~\cite{yu2026visiontrim}
& 56.84
& 51.09
& 63.50
& 82.10
& 77.23
& \textbf{49.33}
& 2248.94
& 78.79
& 82.29\% \\

PruneSiD (ICLR'2026)~\cite{fang2026prunesid}
& 67.12
& 63.92
& 55.70
& 84.58
& 76.72
& 46.33
& 2127.01
& \underline{84.98}
& 84.19\% \\

\hdashline

Dynamic-LLaVA$^\dagger$ (ICLR'2025)~\cite{huang2025dynamicllava}
& 79.21
& 71.73
& 67.40
& 77.00
& 81.87
& 42.56
& 2117.37
& 77.22
& 86.20\% \\

VisionSelector$^\dagger$ (ACM MM'2026)~\cite{zhu2026visionselector}
& \underline{86.40}
& 76.87
& \underline{72.90}
& \textbf{85.42}
& \underline{90.19}
& 48.22
& \underline{2288.56}
& \textbf{85.03}
& \underline{94.50\%} \\

\rowcolor{OursBlue}
\textbf{\method{}}$^\dagger$
& \textbf{90.01}
& \textbf{78.18}
& \textbf{75.60}
& \underline{85.03}
& \textbf{91.39}
& 48.67
& \textbf{2295.87}
& 83.89
& \textbf{95.66\%} \\

\midrule
\multicolumn{10}{c}{%
\cellcolor{gray!8}\textit{Retain 10\% Tokens (90\% Compression Ratio)}} \\
\midrule

FastV (ECCV'2024)~\cite{chen2024fastv}
& 58.64
& \underline{70.83}
& 44.00
& 81.95
& 75.74
& 45.78
& 1940.91
& 65.99
& 77.96\% \\

LLaVA-PruMerge (ICCV'2025)~\cite{shang2025prumerge}
& 42.08
& 56.87
& 41.70
& 81.56
& 71.08
& 45.22
& 1948.58
& 76.52
& 74.10\% \\

VisionZip (CVPR'2025)~\cite{yang2025visionzip}
& 48.29
& 55.94
& 40.40
& 82.60
& 73.09
& 45.44
& 1944.04
& 78.46
& 75.31\% \\

DivPrune (CVPR'2025)~\cite{alvar2025divprune}
& 54.04
& 64.65
& 47.70
& 82.15
& 72.80
& 46.00
& 2015.66
& 79.27
& 79.00\% \\

VisionTrim (ICLR'2026)~\cite{yu2026visiontrim}
& 38.82
& 37.04
& 51.10
& 79.57
& 69.14
& \underline{46.44}
& 2137.21
& 72.12
& 72.24\% \\

PruneSiD (ICLR'2026)~\cite{fang2026prunesid}
& 45.09
& 53.24
& 38.40
& 80.71
& 69.20
& 44.56
& 1953.95
& \textbf{80.13}
& 73.47\% \\

\hdashline

Dynamic-LLaVA$^\dagger$ (ICLR'2025)~\cite{huang2025dynamicllava}
& 61.17
& 68.25
& 55.70
& 75.71
& 76.10
& 43.11
& 1977.58
& 70.80
& 79.05\% \\

VisionSelector$^\dagger$ (ACM MM'2026)~\cite{zhu2026visionselector}
& \underline{72.25}
& 69.68
& \underline{61.90}
& \textbf{84.43}
& \textbf{84.23}
& 45.33
& \underline{2142.62}
& 78.60
& \underline{86.55\%} \\

\rowcolor{OursBlue}
\textbf{\method{}}$^\dagger$
& \textbf{79.32}
& \textbf{73.06}
& \textbf{66.50}
& \underline{83.14}
& \underline{84.00}
& \textbf{47.89}
& \textbf{2158.71}
& \underline{79.90}
& \textbf{89.36\%} \\

\bottomrule
\end{tabular}%
}
\end{table*}

\begin{table*}[t]
\centering
\footnotesize
\caption{Image-language results on Qwen2.5-VL-3B. Full tokens define 100\% retention. Bold and underlined values mark the best and second-best pruning results; $\dagger$ denotes training-based methods that use the same training data.}
\label{tab:comparison_3b}
\setlength{\tabcolsep}{4pt}
\renewcommand{\arraystretch}{0.96}
\resizebox{\textwidth}{!}{%
\begin{tabular}{lccccccccc}
\toprule
\multirow{2}{*}{\textbf{Method}}
& \textbf{DocVQA} & \textbf{TextVQA} & \textbf{OCRBench} & \textbf{ScienceQA} & \textbf{AI2D} & \textbf{MMMU} & \textbf{MME} & \textbf{POPE}
& \multirow{2}{*}{\textbf{Avg.}} \\
& ANLS & Acc. & Acc. & EM & Acc. & Acc. & Score & F1 & \\
\midrule
Qwen2.5-VL-3B
& \textit{92.76} & \textit{77.87} & \textit{78.80} & \textit{80.37} & \textit{90.84} & \textit{46.78} & \textit{2168.46} & \textit{86.94}
& \textit{100.00\%} \\
\midrule
\multicolumn{10}{c}{\cellcolor{gray!8}\textit{Retain 30\% Tokens (70\% Compression Ratio)}} \\
\midrule
FastV (ECCV'2024)~\cite{chen2024fastv}
& 81.66 & 74.02 & 62.90 & 78.63 & 86.40 & \textbf{46.00} & 2048.53 & 82.94
& 93.01\% \\
LLaVA-PruMerge (ICCV'2025)~\cite{shang2025prumerge}
& 65.43 & 64.29 & 55.30 & 78.93 & 79.24 & \textbf{46.00} & 2051.01 & 84.78
& 87.39\% \\
VisionZip (CVPR'2025)~\cite{yang2025visionzip}
& 81.38 & 67.71 & 63.50 & 79.47 & 84.88 & 45.44 & 2065.69 & 86.00
& 92.36\% \\
DART (EMNLP'2025)~\cite{wen2025dart}
& 65.96 & 63.64 & 54.10 & 79.72 & 72.28 & 45.44 & 1996.75 & 84.85
& 85.88\% \\
DivPrune (CVPR'2025)~\cite{alvar2025divprune}
& 73.99 & 70.34 & 63.10 & \textbf{80.42} & 83.42 & 44.78 & 2021.30 & \underline{86.38}
& 91.29\% \\
VisionTrim (ICLR'2026)~\cite{yu2026visiontrim}
& 52.53 & 53.16 & 60.20 & 79.67 & 77.86 & 44.89 & \textbf{2130.78} & 85.68
& 84.86\% \\
PruneSiD (ICLR'2026)~\cite{fang2026prunesid}
& 72.33 & 56.40 & 52.90 & 79.42 & 78.12 & \underline{45.78} & 1979.18 & \textbf{86.51}
& 86.37\% \\
\hdashline
VisionSelector$^\dagger$ (ACM MM'2026)~\cite{zhu2026visionselector}
& \underline{88.02} & \underline{74.80} & \textbf{73.90} & 79.57 & \underline{88.93} & \textbf{46.00} & \underline{2124.92} & 85.60
& \underline{97.05\%} \\
\rowcolor{OursBlue}
\textbf{\method{}}$^\dagger$
& \textbf{90.15} & \textbf{75.90} & \underline{73.60} & \underline{79.82} & \textbf{89.05} & 45.67 & 2118.24 & 84.45
& \textbf{97.23\%} \\
\midrule
\multicolumn{10}{c}{\cellcolor{gray!8}\textit{Retain 20\% Tokens (80\% Compression Ratio)}} \\
\midrule
FastV (ECCV'2024)~\cite{chen2024fastv}
& 72.90 & 71.41 & 56.00 & 78.88 & 83.29 & 45.11 & 1992.58 & 80.01
& 88.94\% \\
LLaVA-PruMerge (ICCV'2025)~\cite{shang2025prumerge}
& 52.58 & 58.44 & 48.40 & 79.97 & 74.06 & 44.89 & 1977.90 & 83.23
& 82.14\% \\
VisionZip (CVPR'2025)~\cite{yang2025visionzip}
& 69.62 & 59.97 & 50.60 & 79.97 & 78.79 & \underline{45.67} & 1892.24 & 84.17
& 85.53\% \\
DART (EMNLP'2025)~\cite{wen2025dart}
& 51.01 & 56.44 & 53.70 & \textbf{80.96} & 69.07 & 45.33 & 1925.04 & 81.93
& 81.54\% \\
DivPrune (CVPR'2025)~\cite{alvar2025divprune}
& 63.82 & 66.76 & 52.80 & 79.87 & 77.59 & 45.00 & 1925.84 & \textbf{85.47}
& 86.21\% \\
VisionTrim (ICLR'2026)~\cite{yu2026visiontrim}
& 41.48 & 45.15 & 54.20 & 77.24 & 73.54 & 44.67 & 2048.03 & 81.62
& 79.04\% \\
PruneSiD (ICLR'2026)~\cite{fang2026prunesid}
& 62.01 & 48.99 & 47.50 & 80.12 & 72.67 & 45.11 & 1955.13 & \underline{85.35}
& 81.81\% \\
\hdashline
VisionSelector$^\dagger$ (ACM MM'2026)~\cite{zhu2026visionselector}
& \underline{82.40} & \textbf{73.71} & \underline{68.50} & 80.07 & \underline{86.95} & 45.33 & \underline{2055.40} & 83.07
& \underline{94.12\%} \\
\rowcolor{OursBlue}
\textbf{\method{}}$^\dagger$
& \textbf{85.09} & \underline{73.51} & \textbf{69.20} & \underline{80.17} & \textbf{87.34} & \textbf{45.89} & \textbf{2071.61} & 83.67
& \textbf{94.96\%} \\
\midrule
\multicolumn{10}{c}{\cellcolor{gray!8}\textit{Retain 10\% Tokens (90\% Compression Ratio)}} \\
\midrule
FastV (ECCV'2024)~\cite{chen2024fastv}
& 54.62 & 65.58 & 41.10 & 79.33 & 77.07 & \underline{45.11} & 1815.70 & 70.28
& 79.98\% \\
LLaVA-PruMerge (ICCV'2025)~\cite{shang2025prumerge}
& 37.51 & 46.43 & 35.00 & 78.93 & 68.43 & 44.56 & 1777.60 & 77.55
& 73.06\% \\
VisionZip (CVPR'2025)~\cite{yang2025visionzip}
& 41.67 & 43.83 & 32.10 & 79.47 & 69.43 & \textbf{45.22} & 1725.49 & 78.60
& 72.99\% \\
DART (EMNLP'2025)~\cite{wen2025dart}
& 32.32 & 44.39 & 31.50 & \underline{79.72} & 65.54 & 44.33 & 1733.77 & 75.41
& 70.58\% \\
DivPrune (CVPR'2025)~\cite{alvar2025divprune}
& 47.13 & 57.80 & 39.20 & 78.04 & 70.14 & 44.78 & 1812.75 & \underline{80.50}
& 77.63\% \\
VisionTrim (ICLR'2026)~\cite{yu2026visiontrim}
& 28.62 & 33.70 & 38.30 & 75.81 & 67.42 & 44.78 & 1840.52 & 77.20
& 70.08\% \\
PruneSiD (ICLR'2026)~\cite{fang2026prunesid}
& 43.20 & 49.28 & 38.20 & 77.94 & 66.45 & \underline{45.11} & 1777.55 & 80.29
& 74.90\% \\
\hdashline
VisionSelector$^\dagger$ (ACM MM'2026)~\cite{zhu2026visionselector}
& \underline{67.12} & \textbf{70.11} & \underline{57.10} & \textbf{80.52} & \underline{80.54} & 44.56 & \underline{1892.52} & \textbf{82.19}
& \underline{87.60\%} \\
\rowcolor{OursBlue}
\textbf{\method{}}$^\dagger$
& \textbf{70.66} & \underline{66.95} & \textbf{58.10} & 79.52 & \textbf{81.35} & \textbf{45.22} & \textbf{1903.15} & \underline{80.50}
& \textbf{87.68\%} \\
\bottomrule
\end{tabular}%
}
\end{table*}

\begin{table*}[t]
\centering
\footnotesize
\caption{Image-language results on LLaVA-OneVision-1.5-8B. Full tokens define 100\% retention. Bold and underlined values mark the best and second-best pruning results; $\dagger$ denotes training-based methods that use the same training data.}
\label{tab:comparison_llava8b}
\setlength{\tabcolsep}{4pt}
\renewcommand{\arraystretch}{0.96}
\resizebox{\textwidth}{!}{%
\begin{tabular}{lccccccccc}
\toprule
\multirow{2}{*}{\textbf{Method}}
& \textbf{DocVQA} & \textbf{TextVQA} & \textbf{OCRBench} & \textbf{ScienceQA} & \textbf{AI2D} & \textbf{MMMU} & \textbf{MME} & \textbf{POPE}
& \multirow{2}{*}{\textbf{Avg.}} \\
& ANLS & Acc. & Acc. & EM & Acc. & Acc. & Score & F1 & \\
\midrule
LLaVA-OV-1.5-8B
& \textit{97.87} & \textit{79.51} & \textit{83.00} & \textit{98.56} & \textit{94.01} & \textit{56.44} & \textit{2271.32} & \textit{88.46}
& \textit{100.00\%} \\
\midrule
\multicolumn{10}{c}{\cellcolor{gray!8}\textit{Retain 30\% Tokens (70\% Compression Ratio)}} \\
\midrule
FastV (ECCV'2024)~\cite{chen2024fastv}
& \underline{86.51} & 72.27 & 59.60 & 91.22 & 83.52 & 53.67 & 2019.84 & 70.41
& 87.01\% \\
DivPrune (CVPR'2025)~\cite{alvar2025divprune}
& 85.23 & \underline{76.03} & \underline{64.50} & \underline{94.65} & \underline{89.35} & \underline{54.11} & \underline{2104.77} & \textbf{88.23}
& \underline{92.47\%} \\
\hdashline
\rowcolor{OursBlue}
\textbf{\method{}}$^\dagger$
& \textbf{94.98} & \textbf{77.50} & \textbf{74.60} & \textbf{97.17} & \textbf{91.45} & \textbf{55.00} & \textbf{2221.99} & \underline{87.00}
& \textbf{96.74\%} \\
\midrule
\multicolumn{10}{c}{\cellcolor{gray!8}\textit{Retain 20\% Tokens (80\% Compression Ratio)}} \\
\midrule
FastV (ECCV'2024)~\cite{chen2024fastv}
& \underline{76.64} & 67.33 & 48.90 & 87.95 & 78.89 & 51.33 & 1918.66 & 67.08
& 80.79\% \\
DivPrune (CVPR'2025)~\cite{alvar2025divprune}
& 75.90 & \underline{72.73} & \underline{55.60} & \underline{92.46} & \underline{84.26} & \underline{52.56} & \underline{2057.83} & \textbf{87.17}
& \underline{87.71\%} \\
\hdashline
\rowcolor{OursBlue}
\textbf{\method{}}$^\dagger$
& \textbf{90.58} & \textbf{75.38} & \textbf{69.90} & \textbf{95.64} & \textbf{89.41} & \textbf{54.56} & \textbf{2178.31} & \underline{86.40}
& \textbf{94.25\%} \\
\midrule
\multicolumn{10}{c}{\cellcolor{gray!8}\textit{Retain 10\% Tokens (90\% Compression Ratio)}} \\
\midrule
FastV (ECCV'2024)~\cite{chen2024fastv}
& 54.43 & 56.47 & 37.40 & 81.51 & 73.15 & 49.56 & 1800.00 & 62.89
& 71.30\% \\
DivPrune (CVPR'2025)~\cite{alvar2025divprune}
& \underline{56.48} & \underline{65.48} & \underline{41.90} & \underline{88.20} & \underline{76.85} & \underline{51.33} & \underline{1897.29} & \textbf{83.92}
& \underline{78.89\%} \\
\hdashline
\rowcolor{OursBlue}
\textbf{\method{}}$^\dagger$
& \textbf{73.67} & \textbf{68.75} & \textbf{51.20} & \textbf{91.08} & \textbf{84.72} & \textbf{53.67} & \textbf{2056.45} & \underline{82.32}
& \textbf{85.58\%} \\
\bottomrule
\end{tabular}%
}
\end{table*}

\subsection{Joint Training and Inference}\label{sec:total-loss}

We combine answer supervision, distribution matching, and consistency across budgets to train the scorer:
\begin{equation}
  \mathcal{L}_{\mathrm{total}}
    = \mathcal{L}_{\mathrm{CE}}
    + \mathcal{L}_{\mathrm{KL}}
    + \lambda(t)\,\mathcal{L}_{\mathrm{MBL}},
\end{equation}
CE and DVD use the softly weighted student path at $b=0.3$. MBL supervises masks at $\{0.3,0.2,0.1\}$ from the same scores, without additional prediction passes at smaller budgets. We increase the MBL weight linearly during training:
\begin{equation}
  \lambda(t)
    = \lambda_{\text{start}}
    + \left(\lambda_{\text{end}} - \lambda_{\text{start}}\right)
      \frac{t}{t_{\max}},
\end{equation}
Here $t$ is the current training step and $t_{\max}$ is the total number of steps. The schedule gradually increases the influence of mask supervision. We use $(\lambda_{\mathrm{start}},\lambda_{\mathrm{end}})=(0.1,2.0)$.

Only scorer parameters are optimized, including the projection weights and, for text-aware scoring, the gate. CE supervises reference answers, DVD matches teacher probabilities, and MBL trains masks across budgets. Confidence decisions and sampled responses are evaluated after training.

At inference, the scorer ranks visual tokens once and retains the top $k=\max(1,\min(N,\operatorname{round}(bN)))$ in their original order. The frozen language model generates from this shortened sequence. The teacher branch and soft masks are unnecessary for selection.

\section{Experiments}\label{sec:experiments}
We evaluate task performance on three backbones and analyze predictive drift and behavior on Qwen2.5-VL-7B. Ablations examine soft teacher targets and multi-budget training.

\subsection{Experimental Setup}\label{sec:exp_setting}

We evaluate Qwen2.5-VL-7B, Qwen2.5-VL-3B, and LLaVA-OneVision-1.5-8B at 30\%, 20\%, and 10\% retention. DPS follows the vision--language projector. The budget is the fraction of visual tokens retained for the frozen language model.

For comparison with VisionSelector~\cite{zhu2026visionselector}, we use the same approximately 144K-sample mixture from Cambrian-737K~\cite{tong2024cambrian}. It contains ChartQA~\cite{masry2022chartqa}, OCR-VQA~\cite{mishra2019ocrvqa}, and a seed-42 10\% sample of the COCO portion~\cite{lin2014microsoft}. Training uses one RTX 5090 and BF16. We use AdamW for one epoch with an effective batch size of 64. The learning rate is $5\times10^{-5}$ with cosine decay and 0.03 warmup, and $d=1792$. The task loss and DVD use $b=0.3$. MBL uses $\{0.3,0.2,0.1\}$. The same trained scorer is used for all evaluated token budgets.

The training-free baselines are FastV~\cite{chen2024fastv}, DART~\cite{wen2025dart}, VisionTrim~\cite{yu2026visiontrim}, VisionZip~\cite{yang2025visionzip}, DivPrune~\cite{alvar2025divprune}, LLaVA-PruMerge~\cite{shang2025prumerge}, and PruneSiD~\cite{fang2026prunesid}. Learnable baselines are Dynamic-LLaVA~\cite{huang2025dynamicllava} and VisionSelector~\cite{zhu2026visionselector}. We use official code when available. Otherwise, we match the backbone, preprocessing, prompts, decoding, and evaluation settings through a shared evaluation interface.

The image benchmarks cover document and scene-text understanding with DocVQA~\cite{mathew2021docvqa}, TextVQA~\cite{singh2019textvqa}, and OCRBench~\cite{liu2024ocrbench}. ScienceQA~\cite{lu2022learn}, AI2D~\cite{kembhavi2016diagram}, and MMMU~\cite{yue2024mmmu} test scientific questions, diagrams, and multidisciplinary understanding. MME~\cite{fu2023mme} and POPE~\cite{li2023pope} add broader multimodal and object-level evaluation. We use \texttt{lmms-eval}~\cite{zhang2024lmmseval} for scoring and MVBench~\cite{li2024mvbench} for video evaluation at 20\% retention.

Tables~\ref{tab:comparison_7b}--\ref{tab:video} report task performance. Section~\ref{sec:common_correct} defines the controls used to measure predictive drift. Section~\ref{sec:threshold_transfer} gives the confidence, sampling, and association protocols alongside the corresponding behavioral results.

\subsection{Task Performance Across Models and Modalities}\label{sec:task_performance}

Table~\ref{tab:comparison_7b} shows that \method{} has the highest average performance retention at every budget. At 10\% retention, its average is 89.36\%. VisionSelector retains 86.55\%, giving \method{} a gain of 2.81 percentage points. At 10\%, the gains over the strongest baselines are especially clear on DocVQA and OCRBench. \method{} also leads MME at 20\% and 10\%, and AI2D at 20\%.

POPE tests object-level yes/no answers. PruneSiD leads F1 at 30\% and 10\%, while VisionSelector leads at 20\%. \method{} achieves the second-highest F1 at 10\%.

\label{sec:cross_backbone}
Tables~\ref{tab:comparison_3b} and~\ref{tab:comparison_llava8b} extend the task evaluation to Qwen2.5-VL-3B and LLaVA-OneVision-1.5-8B.

\method{} leads the average among the methods evaluated on each additional backbone. Its margins over VisionSelector are small on Qwen2.5-VL-3B at 30\% and 10\%. The LLaVA comparison includes FastV and DivPrune, with larger gains under stronger compression on DocVQA and OCRBench. Each backbone uses separately trained scorer weights. The same scorer serves all three retention budgets on that backbone.

The video evaluation provides a further test of retained task performance. On MVBench at 20\% retention, \method{} reaches 66.38\% accuracy compared with 68.10\% for the full model (Table~\ref{tab:video}). It preserves 97.47\% of full-model accuracy. Section~\ref{sec:efficiency} reports the prefill time and peak memory measured in this same video evaluation setting.

\begin{table}[!htb]
\centering
\footnotesize
\caption{MVBench performance and efficiency at 20\% retention. Qwen2.5-VL-7B is used for all methods in this comparison.}
\label{tab:video}
\setlength{\tabcolsep}{3pt}
\renewcommand{\arraystretch}{1.00}
\resizebox{\columnwidth}{!}{%
\begin{tabular}{lccccc}
\toprule
\multirow{2}{*}{\textbf{Method}}
& \textbf{Acc.}
& \textbf{Retention}
& \textbf{Peak Mem.}
& \textbf{Prefill}
& \textbf{Speedup} \\
& (\%) & (\%) & (GB) & (ms) & ($\times$) \\
\midrule
Baseline & 68.10 & 100.00 & 25.97 & 1413.34 & 1.00 \\
FastV    & 65.75 & 96.55  & 24.63 & 851.76  & 1.66 \\
DART     & 65.80 & 96.62  & 18.93 & 832.58  & 1.70 \\
DivPrune & 65.85 & 96.69  & 17.55 & 1184.00 & 1.19 \\
\rowcolor{OursBlue}
\textbf{\method{}}
& \textbf{66.38}
& \textbf{97.47}
& \textbf{16.45}
& \textbf{603.90}
& \textbf{2.34} \\
\bottomrule
\end{tabular}%
}
\end{table}

\subsection{Controlled Analysis of Predictive Drift}\label{sec:common_correct}

We test whether identical greedy answers imply similar response distributions. Since output-matched sets can change with the budget, we also compare fixed samples. Table~\ref{tab:sample_sets} separates method differences at one budget from compression trends measured on the same examples.

\begin{table*}[!t]
\centering
\footnotesize
\caption{Controlled sample sets used in the predictive-drift and behavior analyses. All sets compare the full model and four pruning methods. ``Fixed'' means the same samples are used at all token budgets.}
\label{tab:sample_sets}
\setlength{\tabcolsep}{2pt}
\renewcommand{\arraystretch}{1.14}
\resizebox{\textwidth}{!}{%
\begin{tabular}{@{}lcccl@{}}
\toprule
\textbf{Sample set} & \shortstack{\textbf{Same answer}\\\textbf{required}} & \shortstack{\textbf{Same samples}\\\textbf{across budgets}} & \shortstack{\textbf{Correctness}\\\textbf{required}} & \textbf{Main use} \\
\midrule
Output-matched & \RequiredMark{} & \NotRequiredMark{} & \NotRequiredMark{} & Compare methods at each budget (Table~\ref{tab:strict_fidelity}(A)). \\
Fixed output-matched & \RequiredMark{} & \RequiredMark{} & \NotRequiredMark{} & Follow drift as compression increases (Fig.~\ref{fig:same_output_drift}(c,d)). \\
Fixed correct output-matched & \RequiredMark{} & \RequiredMark{} & \RequiredMark{} & Follow drift on consistently correct samples (Table~\ref{tab:strict_fidelity}(B)). \\
Correct output-matched & \RequiredMark{} & \NotRequiredMark{} & \RequiredMark{} & Relate drift to behavior (Fig.~\ref{fig:drift_failure}; Table~\ref{tab:drift_failure_regression}). \\
Common-correct & \NotRequiredMark{}$^{*}$ & \NotRequiredMark{} & \RequiredMark{} & Confidence controls and sampling tests. \\
\bottomrule
\end{tabular}}
\par\vspace{3pt}
\begin{minipage}{\textwidth}
\footnotesize
\RequiredMark{}: required; \NotRequiredMark{}: not required. $^{*}$All systems answer correctly, but their normalized answer strings may differ.
\end{minipage}
\end{table*}

Let $\hat y(x)$ be the normalized greedy output. Normalization applies Unicode NFKC and case folding and collapses whitespace. For AI2D, it extracts the last standalone A--D option; for TextVQA, it uses the EvalAI answer processor. For each task $d$ and budget $b$, the \emph{output-matched set} contains samples on which the full model and all four pruning methods give the same answer:
\begin{equation}
\mathcal{C}^{\mathrm{same}}_{d,b}=\left\{x:\,
\hat y^{\mathrm{full}}(x)=\hat y^{m,b}(x),\ \forall m\in\mathcal{M}\right\},
\label{eq:same_output}
\end{equation}
where $\mathcal{M}=\{\text{FastV},\text{VisionZip},\text{DivPrune},\text{\method{}}\}$. At a given budget, all methods are compared on the same eligible samples. We measure their agreement with the full-token distribution using JS, KL, and Top-$K$ vocabulary overlap. At each response position, vocabulary overlap is
\begin{equation}
\operatorname{Overlap@}k
=\frac{|\operatorname{TopK}(P_i^{\mathrm{full}})\cap
\operatorname{TopK}(P_i^{m})|}{k}.
\label{eq:topk_overlap}
\end{equation}
At response position $i$, both views are given the same preceding reference tokens $y^*_{<i}$ and use the same tokenizer. For divergence $D$, averaging over response positions gives the sample score:
\begin{equation}
D_x=\frac{1}{L_x}\sum_{i=1}^{L_x}D\!\left(P_i^{\mathrm{full}}(\cdot\mid y^*_{<i}),P_i^{m,b}(\cdot\mid y^*_{<i})\right).
\label{eq:response_position_divergence}
\end{equation}
JS and KL compare next-token probabilities under identical reference prefixes. They measure differences at response positions, not between complete generated sequences. If a greedy answer differs from the reference, its preceding tokens can also differ. Sample scores are averaged within each task, followed by an equally weighted average across DocVQA, TextVQA, OCRBench, and AI2D.

For cross-budget comparisons, we take the intersection of the output-matched sets across 30\%, 20\%, and 10\% retention. This \emph{fixed output-matched set} uses the same samples at all three budgets. The \emph{fixed correct output-matched set} further keeps only samples whose shared answer is correct. Its task average uses AI2D, DocVQA, and TextVQA; OCRBench is excluded because only 20 samples qualify. The common-correct set requires every system to answer correctly at the evaluated budget but allows different normalized answer strings.

\begin{table*}[!t]
\centering
\caption{Predictive fidelity under strict output controls on Qwen2.5-VL-7B. All metrics are equally weighted task averages evaluated under shared reference prefixes. Lower JS/KL and higher vocabulary overlap indicate greater fidelity.}
\label{tab:strict_fidelity}
\footnotesize
\setlength{\tabcolsep}{3pt}
\renewcommand{\arraystretch}{1.00}
\begin{tabular}{@{}p{50pt}*{12}{>{\raggedleft\arraybackslash}p{32.83pt}}@{}}
\toprule
\multicolumn{13}{l}{\textbf{A. Output-matched set} \quad Four tasks; correctness is not required.}\\[2pt]
& \multicolumn{4}{c}{30\% retained ($N=9{,}189$)} & \multicolumn{4}{c}{20\% retained ($N=7{,}201$)} & \multicolumn{4}{c}{10\% retained ($N=4{,}383$)}\\
\cmidrule(lr){2-5}\cmidrule(lr){6-9}\cmidrule(l){10-13}
Method & JS$\downarrow$ & KL$\downarrow$ & Top-5$\uparrow$ & Top-10$\uparrow$ & JS$\downarrow$ & KL$\downarrow$ & Top-5$\uparrow$ & Top-10$\uparrow$ & JS$\downarrow$ & KL$\downarrow$ & Top-5$\uparrow$ & Top-10$\uparrow$\\
\midrule
FastV & 0.0195 & 0.1327 & 90.60 & 90.56 & 0.0234 & 0.1460 & 88.99 & \textbf{89.06} & 0.0171 & 0.0864 & \textbf{87.97} & \textbf{87.13}\\
VisionZip & 0.0162 & 0.1033 & 89.95 & 89.67 & 0.0239 & 0.1457 & 87.76 & 87.32 & 0.0254 & 0.1565 & 83.28 & 82.04\\
DivPrune & 0.0148 & 0.0800 & 88.07 & 87.84 & 0.0202 & 0.1177 & 85.63 & 85.35 & 0.0289 & 0.1512 & 82.42 & 81.51\\
\rowcolor{OursBlue}
\textbf{\method{}} & \textbf{0.0092} & \textbf{0.0612} & \textbf{91.14} & \textbf{90.85} & \textbf{0.0129} & \textbf{0.0782} & \textbf{89.21} & 88.92 & \textbf{0.0127} & \textbf{0.0636} & 86.71 & 85.61\\
\bottomrule
\end{tabular}
\vspace{5pt}

\begin{tabular}{@{}p{50pt}*{6}{>{\raggedleft\arraybackslash}p{71.66pt}}@{}}
\toprule
\multicolumn{7}{l}{B. Fixed correct output-matched set \quad Three tasks; the same $N=3{,}886$ samples at every budget.}\\[2pt]
& \multicolumn{2}{c}{30\% retained} & \multicolumn{2}{c}{20\% retained} & \multicolumn{2}{c}{10\% retained}\\
\cmidrule(lr){2-3}\cmidrule(lr){4-5}\cmidrule(l){6-7}
Method & JS$\downarrow$ & KL$\downarrow$ & JS$\downarrow$ & KL$\downarrow$ & JS$\downarrow$ & KL$\downarrow$\\
\midrule
FastV & 0.0020 & 0.0079 & 0.0033 & 0.0138 & 0.0065 & 0.0298\\
VisionZip & 0.0029 & 0.0132 & 0.0057 & 0.0270 & 0.0177 & 0.1028\\
DivPrune & 0.0042 & 0.0202 & 0.0079 & 0.0371 & 0.0195 & 0.0969\\
\rowcolor{OursBlue}
\method{} & 0.0024 & 0.0109 & 0.0041 & 0.0194 & 0.0100 & 0.0546\\
\bottomrule
\end{tabular}
\par\vspace{3pt}
\begin{minipage}{\textwidth}
\footnotesize
\textit{Notes.} A uses AI2D, DocVQA, OCRBench, and TextVQA; membership varies by budget. Overlap is in percent; bold marks the best value per budget in Panel A.
B uses AI2D (1,632), DocVQA (655), and TextVQA (1,599). OCRBench has only 20 eligible samples and is excluded. Panel B follows within-method trends across retention budgets.
\end{minipage}
\end{table*}

Table~\ref{tab:strict_fidelity}(A) compares methods on shared output-matched samples at each budget. \method{} has the lowest average JS and KL, while FastV leads Top-10 overlap at 20\% and both overlap measures at 10\%. Overlap counts shared high-probability tokens; JS and KL also measure how probability is distributed. Their rankings can therefore differ even when every method generates the same normalized greedy answer.

Table~\ref{tab:strict_fidelity}(B) tests whether stronger compression increases drift when the samples remain fixed and correctly answered. Average JS and KL increase for every method across this three-task cohort. FastV has the lowest drift under this control. Holding both sample membership and correctness fixed reveals a consistent rise in drift as the visual budget decreases.

\begin{figure*}[!t]
\centering\begin{minipage}[t]{.485\textwidth}
\centering
\includegraphics[width=\linewidth]{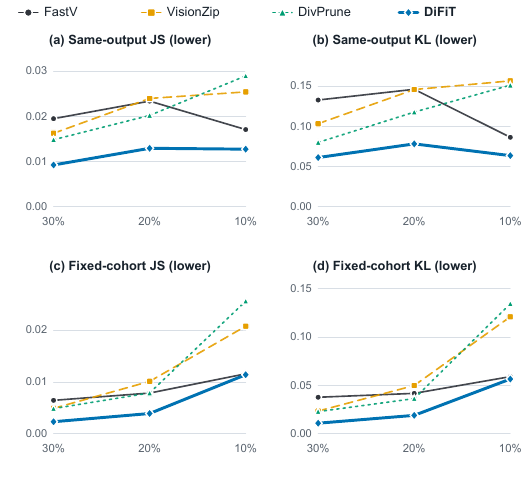}
\caption{Predictive divergence under output matching. Panels (a,b) use four-task output-matched sets containing 9,189, 7,201, and 4,383 samples at 30\%, 20\%, and 10\%. Panels (c,d) track fixed output-matched sets: AI2D (1,665), DocVQA (674), OCRBench (20), and TextVQA (1,755). Points are equally weighted task averages; lower is better. The small OCRBench fixed sample set limits interpretation. The stricter three-task fixed correct output-matched set appears in Table~\ref{tab:strict_fidelity}(B); it is distinct from panels (c,d).}
\label{fig:same_output_drift}
\end{minipage}\hfill\begin{minipage}[t]{.485\textwidth}
\centering
\includegraphics[width=\linewidth]{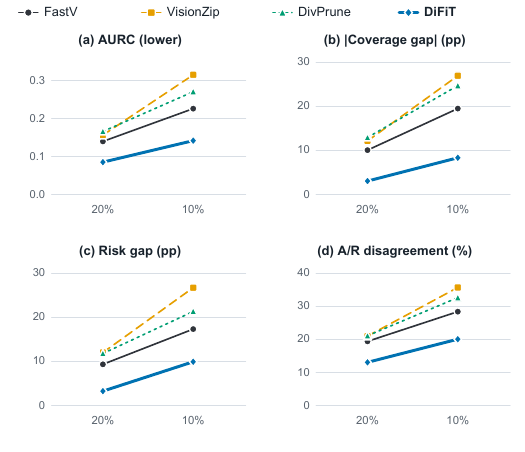}
\caption{Confidence-threshold results on the full set, averaged over five splits and equally across four tasks. AURC evaluates selective prediction across confidence thresholds; the remaining panels transfer a threshold calibrated to 80\% full-model coverage. Coverage gap is plotted by magnitude. Lower is better in all panels, and lines connect the 20\% and 10\% budgets for the same method. \method{} gives the lowest AURC and most closely preserves the full model's decisions at the same confidence threshold.}
\label{fig:threshold_80_slope}
\end{minipage}
\end{figure*}

\begin{figure}[!t]
\centering
\includegraphics[width=\columnwidth]{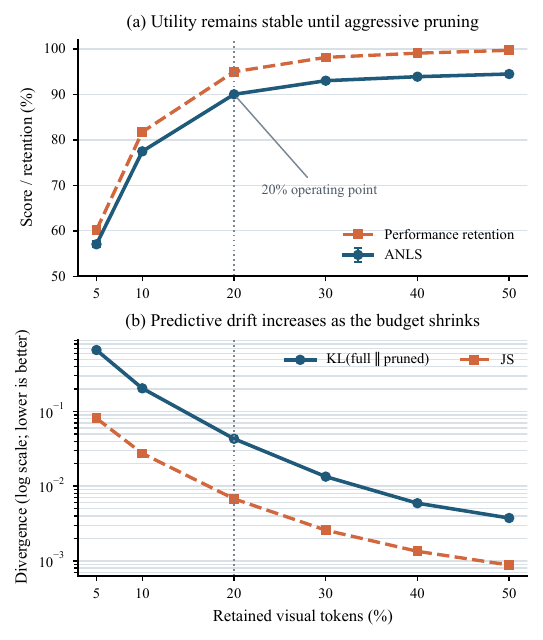}
\caption{Separate DocVQA results at different token budgets on the full evaluation set for \method{}. ANLS includes standard-error bars; performance retention uses the full-token ANLS of 94.78\% as its denominator. Predictive divergences use the full evaluation set without requiring identical answers. In the top panel, dashed denotes retention and solid denotes ANLS.}
\label{fig:e1_docvqa_curve}

\end{figure}

The remaining analyses test how this trend depends on the evaluation set. Figure~\ref{fig:same_output_drift}(c,d) fixes sample membership without requiring correctness and includes OCRBench. Average drift again increases with compression. The separate six-budget DocVQA curve in Fig.~\ref{fig:e1_docvqa_curve} uses the full evaluation set. It shows how task performance and predictive drift change together over a broader budget range, with a sharper decline in task performance once retention falls below 20\%.

These controls separate probability changes from changes in answer text, sample membership, and correctness. Output matching compares methods at a given budget. Fixed membership lets us compare compression levels on the same samples. Requiring correctness further tests whether drift remains when every system already gives the right answer. Drift persists beyond changed or incorrect answers and grows with compression on fixed samples. This reveals differences that task scores alone can miss. We next relate this drift to confidence decisions and stochastic-decoding errors.

\subsection{Behavioral Consistency Beyond Greedy Decoding}\label{sec:threshold_transfer}

\textit{Confidence-threshold transfer.} We first choose a threshold $\tau_d$ for each task to target 90\% full-model coverage. We then reuse it after pruning. Five deterministic splits use 20\% for calibration and 80\% for evaluation (seeds 42, 123, 2026, 3407, and 7819). Each threshold is fixed before held-out evaluation. Confidence is the geometric mean of the $L$ generated-token probabilities, excluding EOS and padding: $c_m(x)=\exp\{L^{-1}\sum_{i=1}^{L}\log p_m(\hat y_i\mid\hat y_{<i},x)\}$. Empty generations have zero confidence. The acceptance decision is
\begin{equation}
A_m(x;\tau_d)=\mathbf{1}\!\left[c_m(x)\geq\tau_d\right].
\label{eq:accept_rule}
\end{equation}
Selective risk is the average error among accepted answers. For tasks with partial-credit scores, error is one minus the normalized task score. AURC is the area under the risk--coverage curve and summarizes selective prediction across thresholds. Coverage gap is the absolute change in the accepted fraction after pruning. Risk gap is the pruned model's selective risk minus the full model's risk. A positive value means more error among accepted answers. A/R disagreement is the fraction of samples whose accept/reject decisions differ.

The full evaluation sets test reuse of the original confidence threshold. Common-correct controls hold correctness fixed, while output-matched controls hold normalized answers fixed regardless of correctness. We also repeat threshold transfer at an 80\% full-model coverage target.

Table~\ref{tab:threshold_transfer} reports full-set results at 90\% target coverage. Pruning changes the number, identity, and error rate of accepted answers. At 10\% retention, \method{} limits the selective-risk increase to 10.86 percentage points, versus 19.59 for FastV. This lower risk gap accompanies fewer changes in which answers pass the reused confidence threshold.

\begin{table}[!htb]
\centering
\footnotesize
\caption{Results from reusing a fixed threshold set for 90\% full-model coverage, averaged over five splits and four tasks. Coverage gap is the absolute pruned-minus-full difference; risk gap remains signed (pp). A/R is accept/reject disagreement. Bold values mark the best result among pruning methods within each token budget.}
\label{tab:threshold_transfer}
\setlength{\tabcolsep}{2.4pt}
\renewcommand{\arraystretch}{1.00}
\resizebox{\columnwidth}{!}{%
\begin{tabular}{lrrrr}
\toprule
\textbf{Method} & \textbf{AURC}$\downarrow$ & $|$\textbf{Cov. Gap}$|\downarrow$ & \textbf{Risk Gap}$\downarrow$ & \textbf{A/R (\%)}$\downarrow$ \\
\midrule
\multicolumn{5}{c}{\cellcolor{gray!8}\textit{Retain 20\% Tokens}} \\
\midrule
FastV & 0.1398 & $7.64$ & $+10.28$ & 13.87 \\
VisionZip & 0.1565 & $8.65$ & $+12.91$ & 14.94 \\
DivPrune & 0.1660 & $8.60$ & $+12.40$ & 14.54 \\
\rowcolor{OursBlue}
\textbf{\method{}} & \textbf{0.0849} & $\mathbf{2.32}$ & $\mathbf{+3.70}$ & \textbf{8.74} \\
\midrule
\multicolumn{5}{c}{\cellcolor{gray!8}\textit{Retain 10\% Tokens}} \\
\midrule
FastV & 0.2264 & $14.88$ & $+19.59$ & 21.04 \\
VisionZip & 0.3159 & $20.88$ & $+28.38$ & 27.02 \\
DivPrune & 0.2712 & $18.15$ & $+22.60$ & 24.19 \\
\rowcolor{OursBlue}
\textbf{\method{}} & \textbf{0.1418} & $\mathbf{6.42}$ & $\mathbf{+10.86}$ & \textbf{14.23} \\
\bottomrule
\end{tabular}%
}
\end{table}

\method{} has the smallest coverage gaps at both budgets. Figure~\ref{fig:threshold_80_slope} shows the same ordering at the 80\% coverage target. The gaps widen as token retention falls from 20\% to 10\%.

\method{} has the lowest disagreement on the common-correct set at both budgets and on the output-matched set at 20\% (Table~\ref{tab:threshold_controls}). FastV is better on the output-matched set at 10\%.

\begin{table}[!htb]
\centering
\footnotesize
\caption{Accept/reject disagreement (\%) at 90\% target coverage on common-correct and output-matched sets, averaged across five splits and four tasks. Lower values indicate closer agreement with the full model's acceptance decisions.}
\label{tab:threshold_controls}
\setlength{\tabcolsep}{3.5pt}
\renewcommand{\arraystretch}{1.00}
\resizebox{\columnwidth}{!}{%
\begin{tabular}{lrrrr}
\toprule
& \multicolumn{2}{c}{\textbf{Common-correct}} & \multicolumn{2}{c}{\textbf{Output-matched}} \\
\cmidrule(lr){2-3}\cmidrule(lr){4-5}
\textbf{Method} & \textbf{20\%} & \textbf{10\%} & \textbf{20\%} & \textbf{10\%} \\
\midrule
FastV & 7.59 & 9.39 & 4.48 & \textbf{5.41} \\
VisionZip & 9.38 & 15.79 & 5.53 & 7.93 \\
DivPrune & 9.27 & 14.48 & 6.68 & 9.10 \\
\rowcolor{OursBlue}
\textbf{\method{}} & \textbf{6.42} & \textbf{9.15} & \textbf{4.29} & 6.40 \\
\bottomrule
\end{tabular}%
}
\end{table}

The 10\% OCRBench output-matched set has only 15--21 examples per split, which limits this comparison. On TextVQA at 10\%, FastV also has lower AURC (0.1474 versus 0.1580) and A/R (12.01\% versus 14.27\%) than \method{}.

\label{sec:stochastic_functional}

\begin{table}[!t]
\centering
\footnotesize
\caption{Four-task average stochastic results; $N$ is the total eligible count. Full and all pruning methods must answer correctly under greedy decoding. Empirical JS is against Full; bold marks the best result among the pruning methods.}
\label{tab:stochastic_functional}
\setlength{\tabcolsep}{2.2pt}
\renewcommand{\arraystretch}{1.00}
\resizebox{\columnwidth}{!}{%
\begin{tabular}{lrrrr}
\toprule
\textbf{Method} & \textbf{Sampled Score}$\uparrow$ & \textbf{Neg. Flip}$\downarrow$ & \textbf{Majority}$\uparrow$ & \textbf{Emp. JS}$\downarrow$ \\
\midrule
\multicolumn{5}{c}{\cellcolor{gray!8}\textit{Retain 20\% Tokens; $N=7{,}141$}} \\
\midrule
Full & 99.05 & 1.31 & 99.75 & -- \\
FastV & 98.07 & 2.40 & 99.32 & 0.07564 \\
VisionZip & 97.65 & 2.89 & 98.96 & 0.09239 \\
DivPrune & 97.03 & 3.75 & 98.89 & 0.09659 \\
\rowcolor{OursBlue}
\textbf{\method{}} & \textbf{98.63} & \textbf{1.81} & \textbf{99.64} & \textbf{0.07494} \\
\midrule
\multicolumn{5}{c}{\cellcolor{gray!8}\textit{Retain 10\% Tokens; $N=4{,}427$}} \\
\midrule
Full & 99.25 & 1.04 & 99.85 & -- \\
FastV & 97.84 & 2.51 & 99.19 & 0.10967 \\
VisionZip & 96.29 & 4.29 & 98.52 & 0.13673 \\
DivPrune & 96.49 & 4.26 & 98.50 & 0.13314 \\
\rowcolor{OursBlue}
\textbf{\method{}} & \textbf{98.56} & \textbf{1.75} & \textbf{99.73} & \textbf{0.10345} \\
\bottomrule
\end{tabular}%
}

\end{table}

\begin{figure}[!htb]
\centering
\includegraphics[width=\columnwidth]{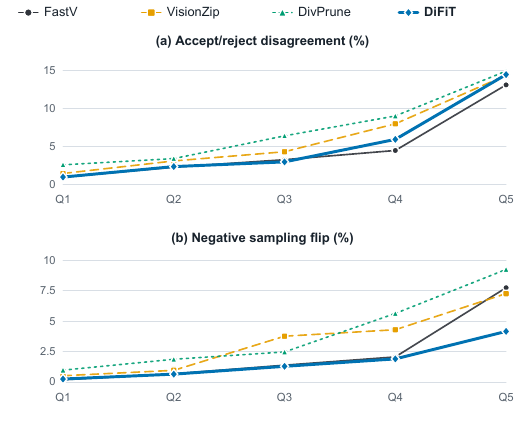}
\caption{Decision disagreements and sampling errors increase with predictive drift on correct, output-matched samples. Five equal-sized groups run from low to high JS within each method, task, and budget. Q1 and Q5 are the lowest- and highest-drift groups. Both rates increase from the lowest- to the highest-drift group for every pruning method.}
\label{fig:drift_failure}
\end{figure}

\textit{Stochastic-decoding stability.} We use each task's common-correct set at each budget. Every evaluated system answers these samples correctly under greedy decoding. We run 10 matched trials with nucleus sampling~\cite{holtzman2020degeneration}, temperature 0.7, and top-$p=0.9$. We use no Top-$K$ truncation, one beam, and repetition penalty 1.0. Methods share deterministic sample--trial seeds. Maximum lengths and stopping rules follow \texttt{lmms-eval}. Table~\ref{tab:stochastic_functional} reports the task-averaged sampled score, negative-flip rate, majority-vote accuracy, and empirical-answer JS.

A \emph{sampling-induced negative flip}, or \emph{negative flip}, occurs when a sampled answer scores below the task maximum despite a correct greedy answer. This includes incorrect responses and answers that receive partial credit.

Sampled score averages task scores across trials, and negative-flip rate counts trials below the maximum score. Majority-vote accuracy scores the most frequent normalized answer. Empirical-answer JS compares normalized-answer frequencies from 10 trials with Full. It describes observed answers. Response-position JS instead compares next-token probabilities at matched positions under shared prefixes.

Table~\ref{tab:stochastic_functional} tests whether correct greedy answers remain correct under sampling. At 10\% retention, \method{} reduces the negative-flip rate from FastV's 2.51\% to 1.75\%, a 30.3\% relative reduction. This comparison uses four-task subsets that all evaluated systems answer correctly under greedy decoding. \method{} has fewer negative flips than FastV at both budgets, but its rates still exceed those of the full-token model.

\method{} also leads the pruning methods in sampled score, majority-vote accuracy, and empirical-answer JS. Pruning can therefore turn correct greedy answers into errors under sampling. The empirical-answer JS values at 20\% are close: 0.07494 for \method{} and 0.07564 for FastV.

\begin{table}[!b]
\centering
\footnotesize
\caption{Association of JS drift with decision disagreements and sampling errors after adjustment. Entries are odds ratios per one-SD increase in JS, with 95\% cluster-robust CIs. Models control full-model confidence, prediction margin, task, and budget. The A/R analysis has 15,436 observations; the sampling analysis has 86,680 trials from 8,668 observations. Samples can recur across budgets.}
\label{tab:drift_failure_regression}
\setlength{\tabcolsep}{3.4pt}
\renewcommand{\arraystretch}{1.05}
\resizebox{\columnwidth}{!}{%
\begin{tabular}{lcc}
\toprule
\textbf{Method} & \textbf{A/R OR [95\% CI]} & \textbf{Negative-flip OR [95\% CI]} \\
\midrule
\method{} & 1.283 [1.202, 1.368] & 1.335 [1.250, 1.426] \\
FastV & 1.589 [1.478, 1.708] & 1.547 [1.457, 1.642] \\
VisionZip & 1.357 [1.288, 1.430] & 1.380 [1.314, 1.450] \\
DivPrune & 1.373 [1.306, 1.443] & 1.389 [1.327, 1.455] \\
\bottomrule
\end{tabular}%
}
\end{table}

\textit{Drift--behavior association.} We construct the correct output-matched set in Table~\ref{tab:sample_sets} separately for each task and budget. Every system must give the same normalized greedy answer and receive full credit. Samples need not qualify at other budgets. Each analysis uses held-out samples with the required drift and behavioral measurements. This fixes answer identity and correctness while allowing response probabilities to vary.

We group these samples by response-position JS. Within each method, task, and budget, we form five equal-sized groups and measure disagreement and negative flips. Rates are averaged equally across nonempty task and budget groups.

Logistic regressions use JS standardized within each method and control for full-model confidence, first-token prediction margin, task, and budget. The margin is the gap between the two largest first-token probabilities. Standard errors are clustered by sample to account for repeated trials and budgets.

On correct, output-matched samples, disagreement and negative flips increase from the lowest- to the highest-drift group for every pruning method (Fig.~\ref{fig:drift_failure}).

We adjust for confidence, prediction margin, task, and budget. Higher JS remains associated with more disagreement and negative flips for every method (Table~\ref{tab:drift_failure_regression}). All 95\% confidence intervals lie above one. A one-standard-deviation increase in JS corresponds to higher event odds within each method. The adjusted results agree with the trends in Fig.~\ref{fig:drift_failure}.

Together, Table~\ref{tab:strict_fidelity}, Fig.~\ref{fig:drift_failure}, and Table~\ref{tab:drift_failure_regression} characterize predictive drift beyond greedy-answer agreement and its association with confidence decisions and sampling failures.

\begin{table}[!b]
\centering
\footnotesize
\caption{Hyperparameter sensitivity on Qwen2.5-VL-7B at 20\% token retention. Each variant changes one setting from $\lambda:0.1$--$2.0$, $\gamma=0.5$, $\tau=2$. Avg. Ret. averages the five scores normalized by Full in Table~\ref{tab:comparison_7b}. The last row uses our scores from that table.}
\label{tab:hyperparameters}
\setlength{\tabcolsep}{3pt}
\begin{tabular}{@{}lrrrrrr@{}}
\toprule
Config & DocVQA & OCRBench & AI2D & MME & POPE & Avg. \\
 & ANLS & Acc. & Acc. & Score & F1 & Ret. \\
\midrule
$\lambda$: $0.1$--$1.0$ & 90.17 & 74.40 & 89.51 & 2287.93 & 81.47 & 94.43\% \\
$\lambda$: $0.1$--$3.0$ & \textbf{90.61} & 76.10 & 90.41 & 2262.23 & 82.41 & 95.12\% \\
$\gamma=0$ & 89.65 & \textbf{76.80} & 91.03 & 2283.25 & 82.80 & 95.49\% \\
$\gamma=1.0$ & 88.26 & 69.60 & 88.44 & 2233.91 & \textbf{83.91} & 92.75\% \\
$\tau=1$ & 90.03 & 73.60 & 90.19 & 2267.36 & 83.22 & 94.58\% \\
$\tau=3$ & 89.99 & 74.80 & 89.90 & 2276.77 & 82.97 & 94.82\% \\
\rowcolor{OursBlue}
\textbf{\method{}} & 90.01 & 75.60 & \textbf{91.39} & \textbf{2295.87} & 83.89 & \textbf{95.71\%} \\
\bottomrule
\end{tabular}
\end{table}

\begin{table}[!b]
\centering
\footnotesize
\caption{Training-data mixtures at 20\% retention from earlier independent checkpoints, distinct from Table~\ref{tab:comparison_7b}. Ratios denote ChartQA/OCR-VQA/COCO.}
\label{tab:training_data_ablation}
\setlength{\tabcolsep}{2pt}
\renewcommand{\arraystretch}{1.00}
\resizebox{\columnwidth}{!}{%
\begin{tabular}{ccccc}
\toprule
\textbf{Data Ratio} & \textbf{Time} & \textbf{DocVQA} & \textbf{OCRBench} & \textbf{MME} \\
\midrule
10 / 10 / 1   & 40m & 88.09 & 73.40 & 2264.89 \\
100 / -- / -- & 50m & 84.52 & 64.70 & 2169.48 \\
50 / 50 / --  & 1h40m & 89.37 & 75.00 & 2291.62 \\
-- / 100 / -- & 2h50m & 87.65 & 74.70 & 2287.69 \\
\midrule
\rowcolor{OursBlue}
\textbf{100 / 100 / 10} & 5h30m & \textbf{90.17} & \textbf{76.10} & \textbf{2295.87} \\
\bottomrule
\end{tabular}%
}
\end{table}

\subsection{Computational Efficiency}\label{sec:efficiency}

\begin{figure}[!t]
\centering
\includegraphics[width=0.98\columnwidth]{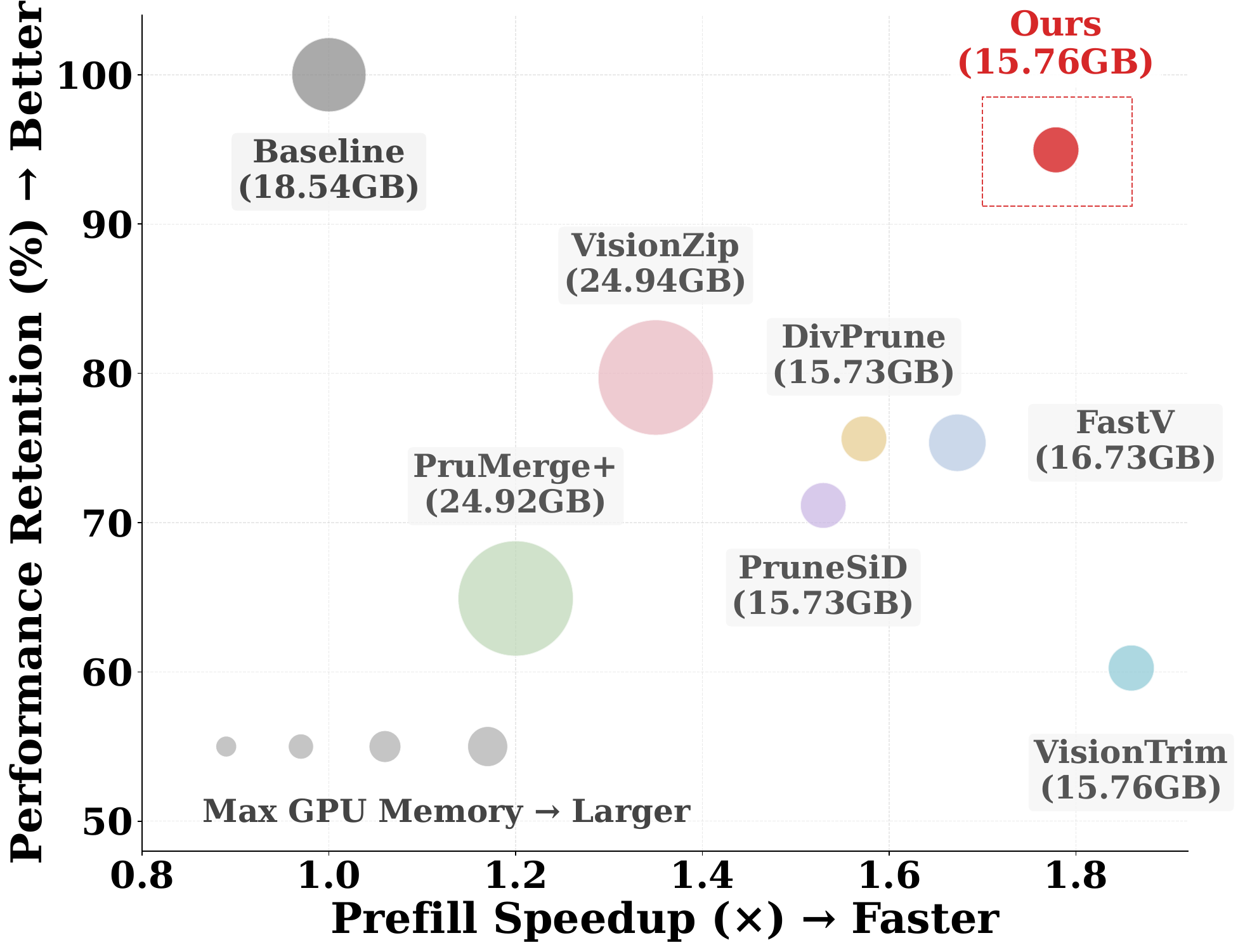}
\caption{DocVQA performance--efficiency trade-off. The horizontal axis shows prefill speedup relative to the full-token model, and the vertical axis shows performance retention. Bubble sizes and annotations indicate peak GPU memory. Token-selection overhead can increase peak memory despite a smaller retained sequence.}
\label{fig:efficiency}
\end{figure}

Figure~\ref{fig:efficiency} compares DocVQA performance retention, prefill speedup, and peak GPU memory. \method{} balances performance and speed, although some methods use slightly less memory. Selection costs computation and temporary storage, so reducing token count need not reduce latency and memory proportionally.

On MVBench at 20\% retention, \method{} lowers prefill time from 1413.34 to 603.90\,ms, a 2.34$\times$ speedup (Table~\ref{tab:video}). Prefill measures processing before response decoding. Peak memory falls from 25.97 to 16.45\,GB, the lowest among the compared methods, while accuracy reaches 66.38\%. These measurements compare task quality, prefill latency, and peak memory at the same token budget and video-evaluation setting.

\begin{table*}[!t]
\centering
\footnotesize
\caption{Comparison of training losses at 20\% retention. Task columns use each benchmark's standard score under greedy decoding. Macro retention averages each task score divided by its full-token reference. KL, JS, and overlap are averages across tasks on the full evaluation sets. All variants use identical data and optimization settings.}
\label{tab:ablation_component}
\setlength{\tabcolsep}{3.6pt}
\renewcommand{\arraystretch}{1.05}
\resizebox{\textwidth}{!}{%
\begin{tabular}{@{}lccccccccc@{}}
\toprule
\textbf{Variant} & \textbf{AI2D} & \textbf{DocVQA} & \textbf{OCRBench} & \textbf{TextVQA} & \textbf{Macro Ret.} & \textbf{Macro KL $\downarrow$} & \textbf{Macro JS $\downarrow$} & \textbf{Top-5 $\uparrow$} & \textbf{Top-10 $\uparrow$} \\
\midrule
Hard-teacher distillation & 90.97 & \textbf{90.49} & 75.30 & 75.37 & 93.24\% & 0.1785 & 0.0263 & 86.15\% & 85.67\% \\
CE + MBL (w/o DVD) & 89.93 & 89.47 & 73.80 & 71.39 & 91.06\% & 0.2014 & 0.0298 & 85.65\% & 85.08\% \\
w/o cross-budget MBL (single-budget) & 89.64 & 89.13 & 74.10 & 75.55 & 92.22\% & 0.1690 & 0.0259 & 87.26\% & 86.99\% \\
\rowcolor{OursBlue}
\textbf{\method{} (soft DVD + MBL)} & \textbf{91.39} & 90.01 & \textbf{75.60} & \textbf{78.18} & \textbf{94.16\%} & \textbf{0.1609} & \textbf{0.0233} & \textbf{87.57\%} & \textbf{87.16\%} \\
\bottomrule
\end{tabular}%
}

\end{table*}

\begin{figure*}[!t]
    \centering
    \includegraphics[width=0.93\textwidth]{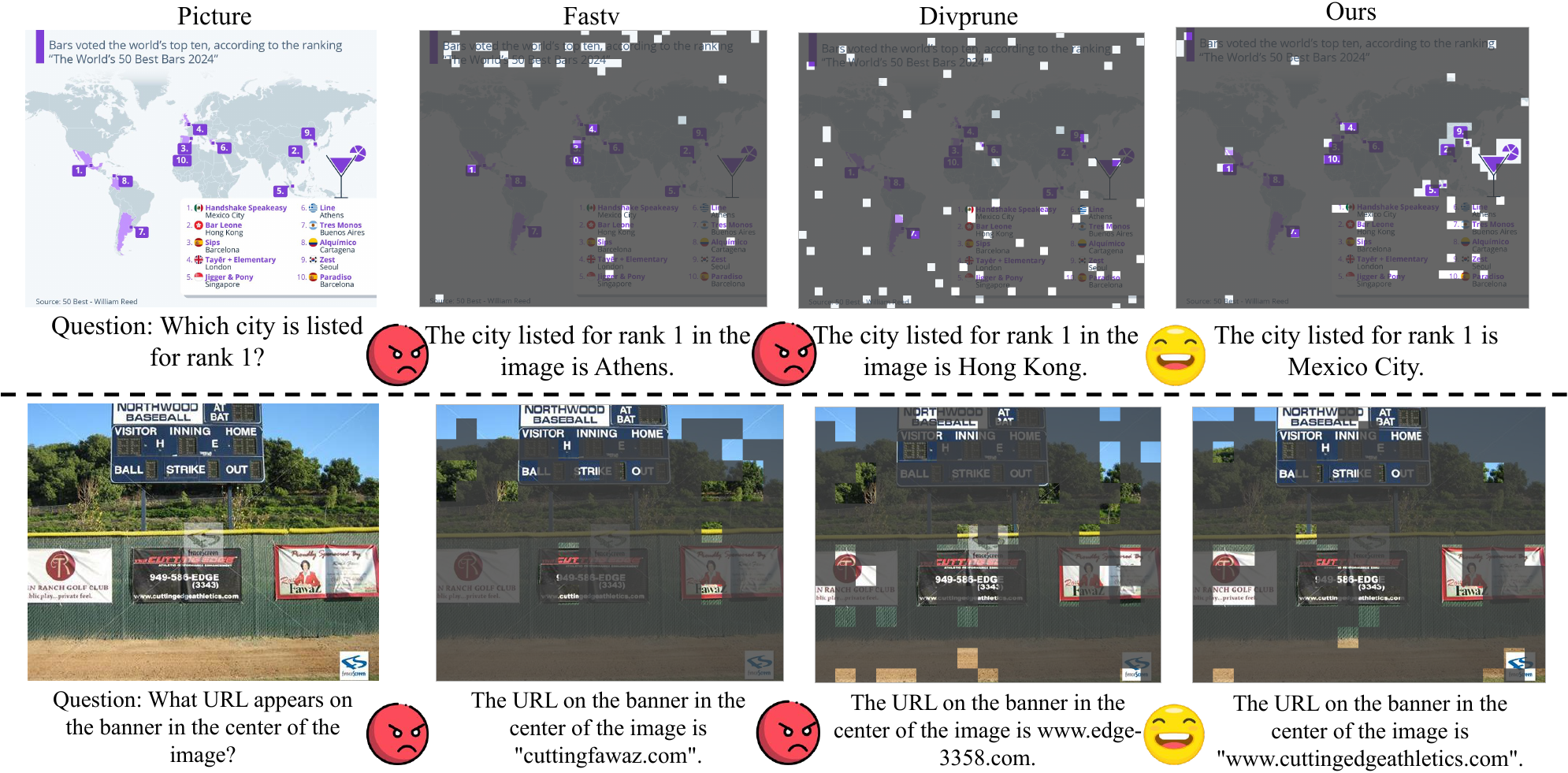}
    \caption{Qualitative token selections for a dense infographic at 10\% retention (top) and scene text at 20\% (bottom). Columns show the input image and retained regions for FastV, DivPrune, and \method{}, with generated answers below each selection.}
    \label{fig:qualitative}

\end{figure*}

\subsection{Ablation and Selection Analysis}\label{sec:ablation}

We test how soft teacher targets and multi-budget training affect task retention and predictive fidelity (Table~\ref{tab:ablation_component}). All four variants use the same ChartQA, OCR-VQA, and COCO mixture. They share micro-batch 8, gradient accumulation 8, effective batch 64, and ZeRO-3 without offload. They differ only in the training objective. The hard-teacher variant replaces soft probabilities with the teacher's most likely token at each supervised response position. Gradients stop at the teacher:
\begin{equation}
\mathcal L_{\mathrm{hard}}=
-\frac{1}{\sum_i m_i}\sum_i m_i
\log P_i^{\mathrm{soft}}\!\left(\arg\max_v P_i^{\mathrm{full}}(v)\right),
\end{equation}
using student probabilities without temperature scaling. The hard-teacher variant retains the task and MBL objectives. The CE + MBL variant removes teacher distillation and retains mask supervision. Soft targets supervise the probabilities of alternative tokens. Hard targets supervise only the most likely teacher token at each response position.

Soft DVD leads average performance retention and vocabulary overlap and has the lowest KL/JS (Table~\ref{tab:ablation_component}). The hard teacher leads only on DocVQA. Compared with CE + MBL, soft DVD raises macro retention from 91.06\% to 94.16\% and lowers macro JS from 0.0298 to 0.0233. Full teacher probabilities thus improve both task retention and fidelity here. Removing cross-budget MBL lowers retention and raises KL/JS, supporting mask supervision across budgets alongside distribution matching.

The default configuration leads the five-task average in Table~\ref{tab:hyperparameters}. Using only hard targets at the smaller budgets ($\gamma=1$) causes the largest drop, especially on OCRBench. Using only larger-budget guidance at the smaller budgets ($\gamma=0$) also lowers the average; the largest-budget hard target remains. Neither the tested MBL schedules nor temperatures improve the average. Individual tasks favor different settings: a larger MBL endpoint leads DocVQA, while $\gamma=0$ leads OCRBench. These differences favor assessing settings across tasks.

Table~\ref{tab:training_data_ablation} compares earlier independent checkpoints with different training-data compositions and amounts. The 100/100/10 mixture leads all three benchmarks and takes 5h30m to train. The 50/50 mixture takes 1h40m and scores closer to the largest mixture than either single-source run. Using ChartQA and OCR-VQA together balances training time and task performance.

\label{sec:qualitative}
Figure~\ref{fig:qualitative} shows query-relevant selection. At 10\% retention, \method{} links the rank to ``Mexico City.'' FastV and DivPrune miss this association. At 20\%, it retains the center-banner URL, whereas the baselines keep incomplete or neighboring text. Both examples require preserving the relation between the requested detail and its surrounding visual context.

\subsection{Limitations}\label{sec:discussion}

Hard selection discards entire tokens. Evidence can be lost when the retained tokens do not represent the discarded regions. The frozen visual encoder cannot adapt its features to a smaller token budget. This limits the model's ability to recover details spread across many visual locations. Local aggregation or compression-aware visual adaptation could help retained tokens carry information from discarded regions.

Shared reference prefixes keep response history identical in the full and pruned views. During free generation, an early token change alters the context for later tokens. Errors can accumulate as the generated histories diverge. Response-position matching under shared reference prefixes does not capture this accumulation. Tests with longer generated responses and a wider range of backbones would extend the four-task behavioral evaluation. They would also help clarify how errors accumulated during generation affect confidence decisions and the answers obtained through sampling.

\section{Conclusion}\label{sec:conclusion}

Visual token pruning can preserve greedy answers while changing the probabilities behind them. On fixed samples with identical correct answers, average drift increases as compression grows. Separate analyses link drift to changes in confidence decisions and sampled answers. \method{} learns token selection from full-token probabilities and reference answers while keeping the MLLM frozen. One ranking supports multiple budgets and improves task retention and behavioral consistency on the evaluated benchmarks. These findings support assessing pruning through task quality, predictive fidelity, confidence decisions, sampled responses, and measured inference cost.

\clearpage
\begingroup
\setlength{\emergencystretch}{0pt}
\bibliographystyle{IEEEtran}
\bibliography{IEEEtranBSTCTL,references}
\endgroup

\begingroup
\makeatletter
\def\@IEEEBIOphotowidth{0.72in}
\def\@IEEEBIOphotodepth{0.9in}
\def\@IEEEBIOhangwidth{0.84in}
\def\@IEEEBIOhangdepth{0.9in}
\def\@IEEEBIOskipN{0.75\baselineskip}
\makeatother


\begin{IEEEbiography}[{\includegraphics[width=0.72in,height=0.9in,clip,keepaspectratio]{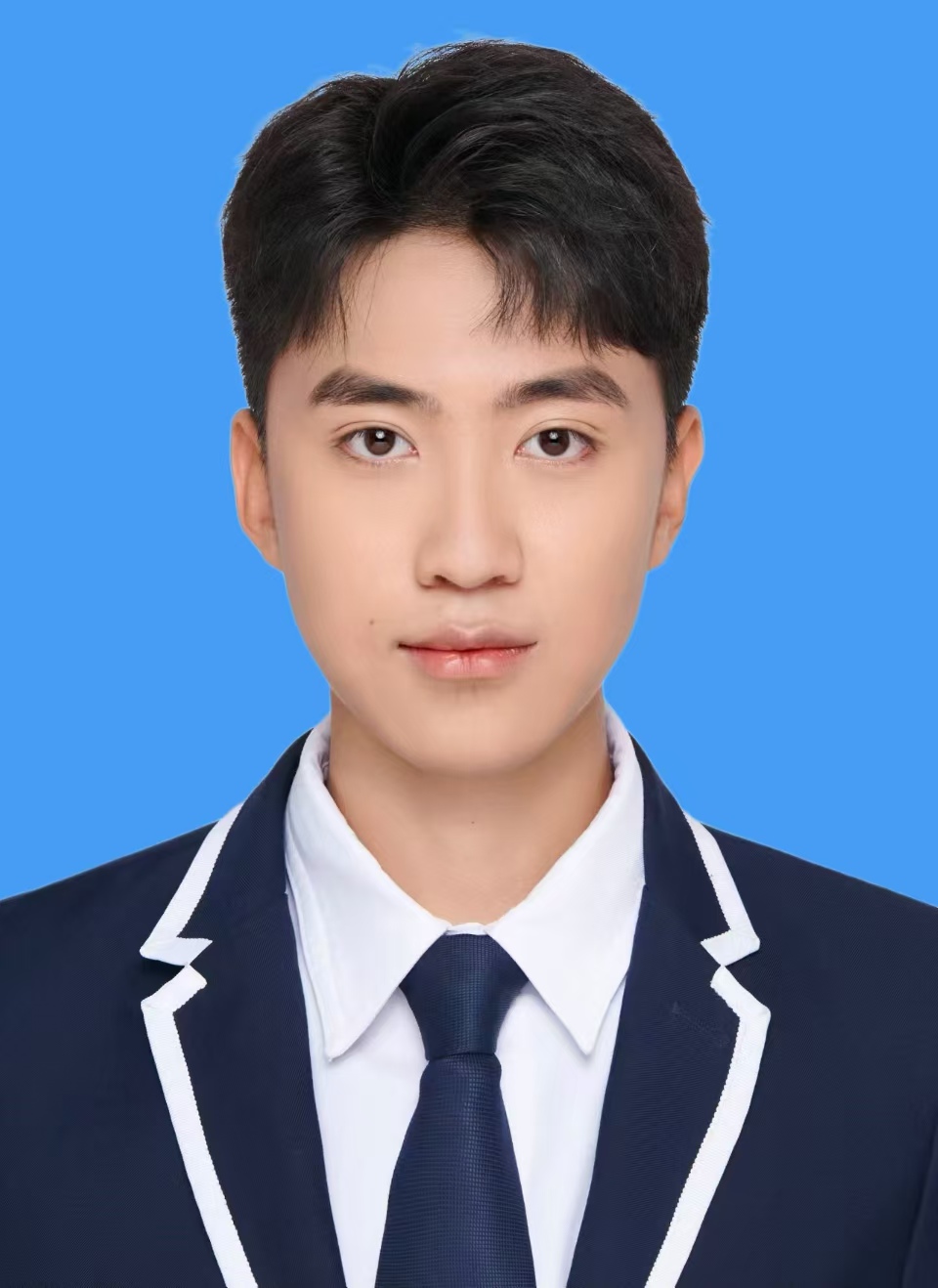}}]{Jingyuan Zhang}
is currently pursuing the B.Eng. degree in robotics engineering with the School of Artificial Intelligence, Anhui University, Hefei, China. His research interests include multimodal large language models, visual token compression, efficient inference, and zero-shot visual anomaly detection. He has published research at ACM MM.
\end{IEEEbiography}

\begin{IEEEbiography}[{\includegraphics[width=0.72in,height=0.9in,clip,keepaspectratio]{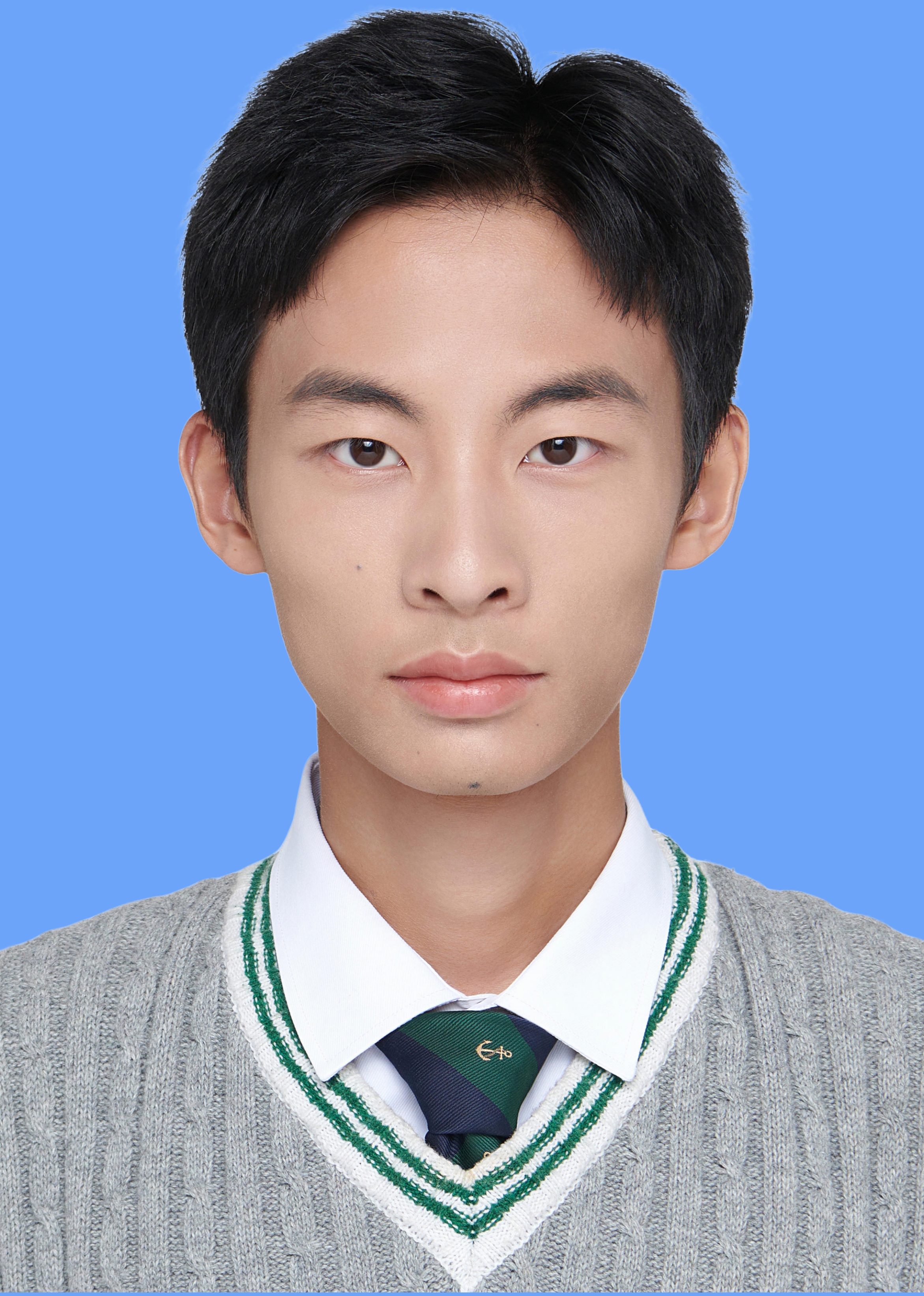}}]{Nihao Niu}
is currently pursuing the B.Eng. degree in artificial intelligence with the School of Artificial Intelligence, Anhui University, Hefei, China. His research interests include multimodal large language models, visual token compression, efficient deep learning, and neural network pruning.
\end{IEEEbiography}

\begin{IEEEbiography}[{\includegraphics[trim=176.280728bp 0bp 176.280728bp 0bp,width=0.65553869in,height=0.9in,clip,keepaspectratio]{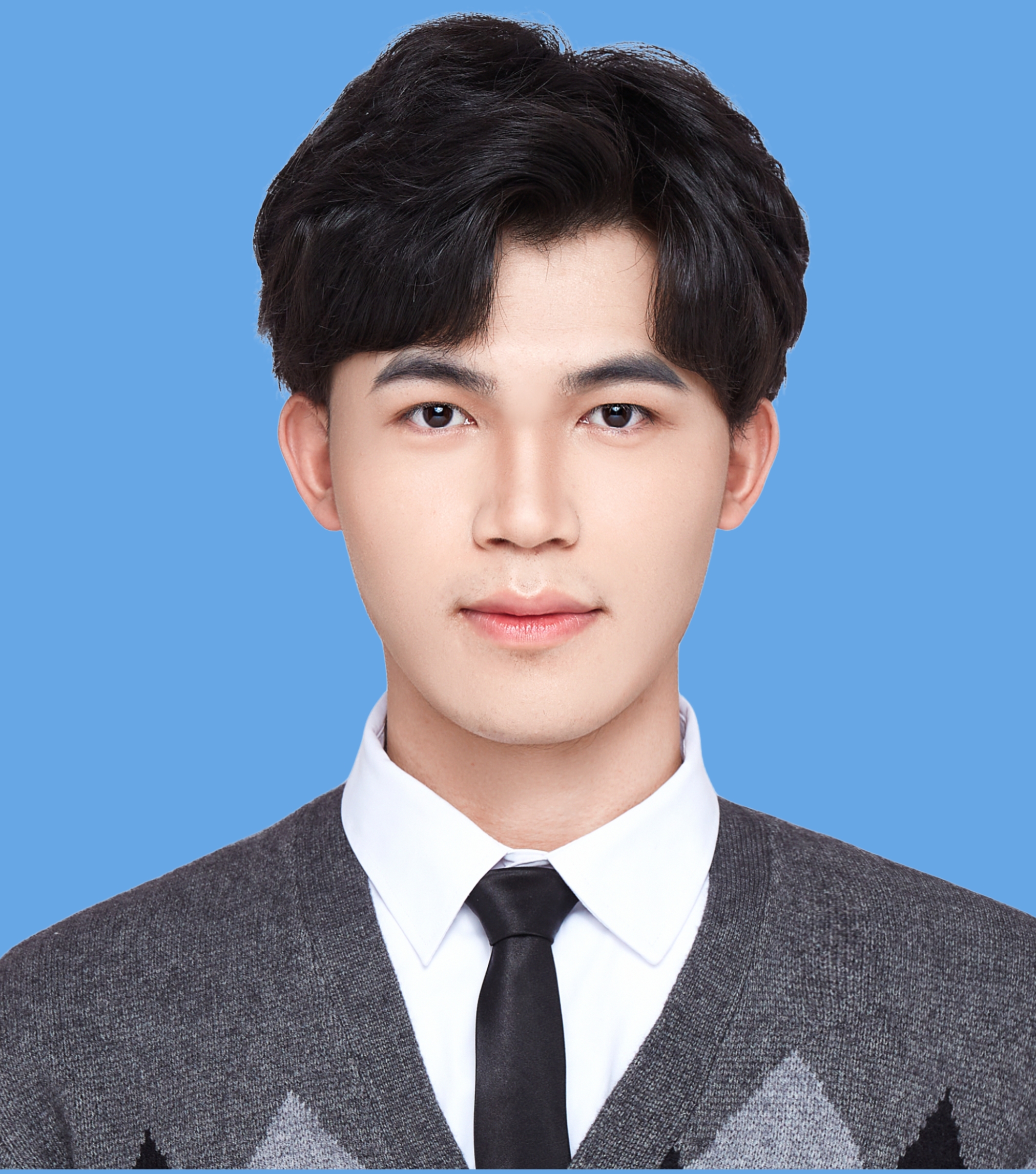}}]{Xinyang Lv}
is currently pursuing the M.S. degree in computer technology at Anhui University, Hefei, China. His research interests include vision-language models, remote sensing image understanding, multimodal learning, and model compression, with a particular focus on efficient multimodal models and visual token pruning.
\end{IEEEbiography}

\begin{IEEEbiography}[{\includegraphics[
  trim=0bp 34.117708bp 0bp 34.117708bp,
  width=0.65553869in,
  height=0.9in,
  clip,
  keepaspectratio
]{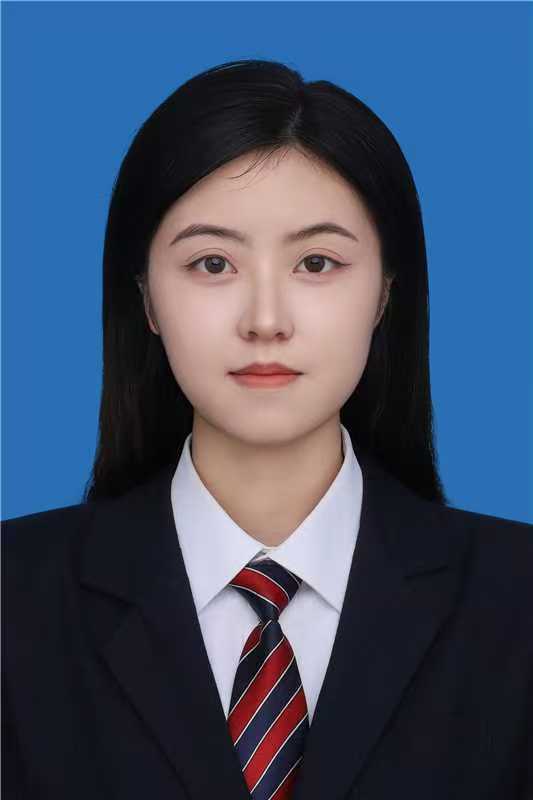}}]{Beibei Han}
is currently pursuing the M.S. degree with the School of Artificial Intelligence, Anhui University, Hefei, China. She received the B.S. degree from Anhui University of Science and Technology, China, in 2025. Her research interests include computer vision and token pruning. She has published research at ACM MM.
\end{IEEEbiography}

\begin{IEEEbiography}
[{\includegraphics[width=0.72in,height=0.9in,clip,keepaspectratio]
{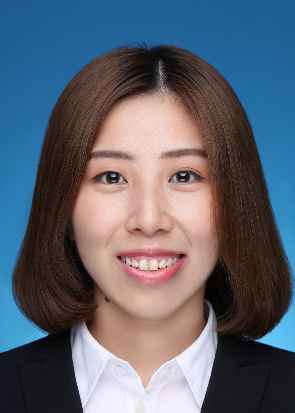}}]{Xiaoyun Wang} is a Ph.D student at the School of Artificial Intelligence, Anhui University. She received the B.S. degree from the Yanshan University, Qinhuangdao, China, in 2016,  and the M.S. degree from Beijing Jiaotong University, Beijing, China, in 2019. Her current research interests include computer vision, neural network compression, high performance computing.
\end{IEEEbiography}

\begin{IEEEbiography}[{\includegraphics[width=0.72in,height=0.9in,clip,keepaspectratio]{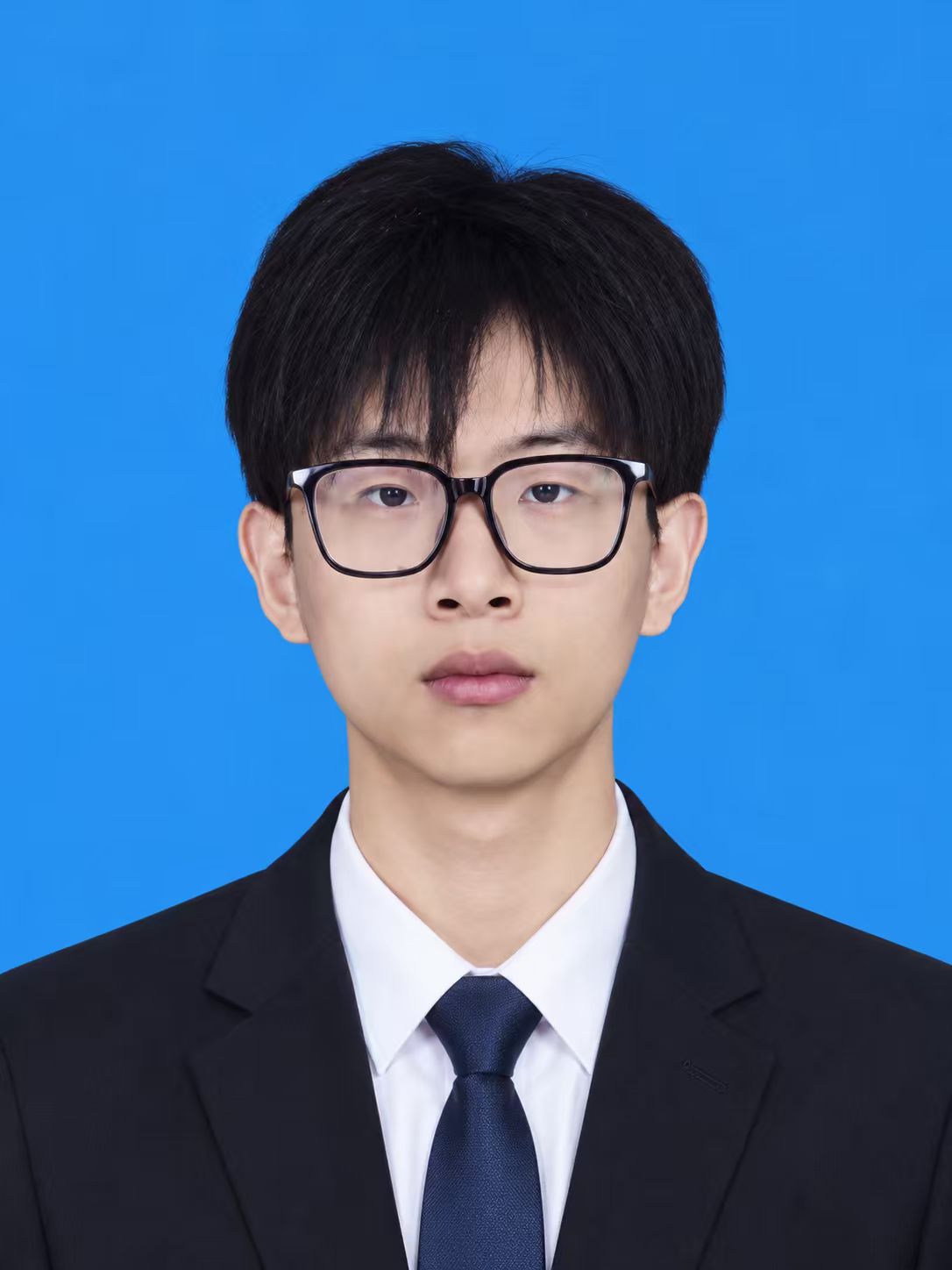}}]{Yanning Hou}
received the bachelor's degree from Anhui University, Hefei, China, in 2024. He is currently pursuing the M.S. degree with the College of Intelligence Science and Technology, National University of Defense Technology (NUDT), Changsha, China. His research interests include GraphRAG, retrieval-augmented generation, intelligent agents, efficient AI systems, and industrial anomaly detection. His recent work focuses on structured knowledge representation, retrieval, and reasoning for efficient and reliable large language model agents, as well as zero-shot visual anomaly detection. He has published research at IJCAI and CVPR.
\end{IEEEbiography}

\begin{IEEEbiography}[{\includegraphics[width=0.72in,height=0.9in,clip,keepaspectratio]{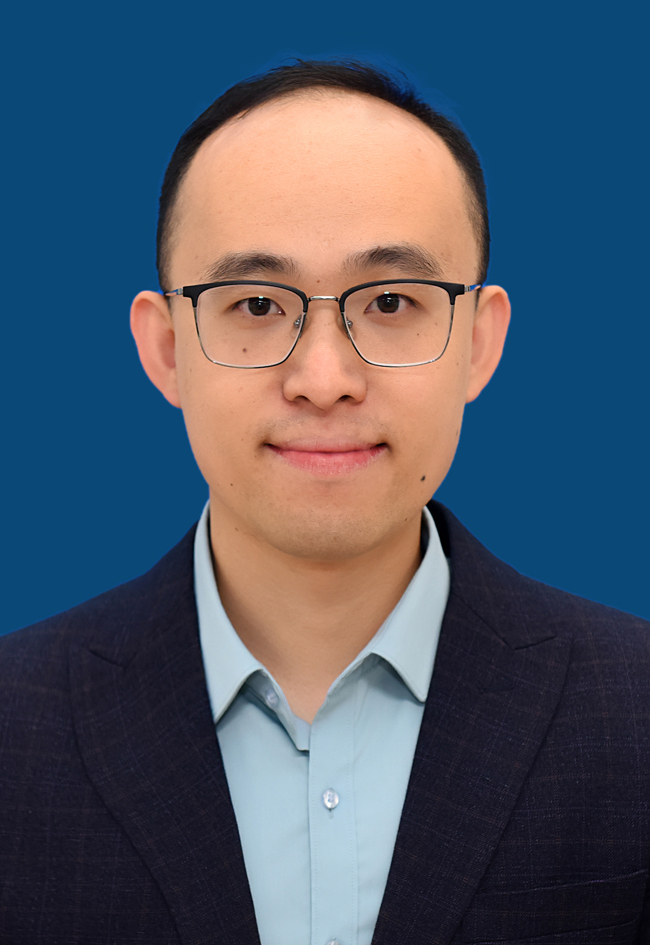}}]{Ke Xu}  (Member, IEEE) is an Associate Professor at the State Key Laboratory of Opto-Electronic Information Acquisition and Protection Technology and School of Artificial Intelligence, Anhui University, Hefei, China. He received the B.S. degree in Hefei University of Technology, China, in 2016, and the Ph.D degree in Institute of Information Science, Beijing Jiaotong University, Beijing, China, in 2021. His research interests include neural network compression, high performance computing architectures for embedded applications and computer vision.
\end{IEEEbiography}


\vfill

\endgroup

\end{document}